\documentclass[lettersize,journal]{IEEEtran}
\usepackage{amsmath,amsfonts}
\usepackage{algorithm,algpseudocode}

\usepackage[caption=false,font=normalsize,labelfont=sf,textfont=sf]{subfig}
\usepackage{textcomp}
\usepackage{stfloats}
\usepackage{multirow}
\usepackage{url}
\usepackage{adjustbox}
\usepackage{verbatim}
\usepackage{graphicx, multicol}
\usepackage{cite}
\usepackage{array}
\usepackage{epstopdf}
\usepackage{color}
\usepackage{enumerate}
\usepackage{float}
\usepackage{booktabs}
\usepackage{siunitx}
\usepackage{subfig}
\usepackage{hyperref}
\usepackage{arydshln}
\usepackage{tikz}
\usepackage{balance} 
\usepackage{pifont}
\usepackage{wrapfig}

\makeatletter
\newcommand{\closetagspacing}{\renewcommand{\tag@spacing}{0pt}}
\makeatother
\algrenewcommand\algorithmicindent{0.6em}%
\hypersetup{
	colorlinks=False,
	linkcolor=cyan,
	filecolor=blue,
	urlcolor=red,
	citecolor=green,
}
\newcommand\highlightReference[1]{%
  \expandafter\newcommand\csname highlightReference-#1\endcsname{}%
}
\let\oldbibitem\bibitem
\def\bibitem#1 #2\par{%
  \expandafter\ifx\csname highlightReference-#1\endcsname\relax
    \oldbibitem{#1}#2\par
  \else
    \oldbibitem{#1}\highlight{#2}\par
  \fi
}

\newcommand\highlight[1]{\textcolor{red}{#1}}

\begin{document}
\title{Towards Adaptive Open-Set Object Detection via Category-Level Collaboration Knowledge Mining}

\author{Yuqi Ji,
        Junjie Ke,
        Lihuo He,~\IEEEmembership{Member,~IEEE},
        Lizhi Wang, ~\IEEEmembership{Member,~IEEE},
        Xinbo Gao,~\IEEEmembership{Fellow,~IEEE}
\thanks{This work was supported by the New Generation Artificial Intelligence-National Science and Technology Major Project (2025ZD0123601) and the National Natural Science Foundation of China (Grant No. 62276203). \emph{(Corresponding author: Lihuo He}.)}
\thanks{Yuqi Ji, Junjie Ke, Lihuo He and Xinbo Gao are with the School of Electronic Engineering, Xidian University, Xi'an 710071, China, and also with the Interdisciplinary Institute of Artificial Intelligence, Faculty of Infor-X, Xidian University, Xi'an, Shaanxi 710126, China. Yuqi Ji and Junjie Ke contributed equally. (e-mail: lhhe@mail.xidian.edu.cn.)}
\thanks{Lizhi Wang is with the School of Artificial Intelligence, Beijing Normal University, Beijing 100875, China}
}



\maketitle

\begin{abstract}
Existing object detection methods struggle to generalize across increasingly data domains while simultaneously adapting to the emergence of novel categories. To tackle this challenge, adaptive open-set object detection (AOOD) has been introduced, which employs supervised training on base categories within the source domain while enabling unsupervised adaptation to both base and novel categories in the target domain. However, existing AOOD approaches are still hindered by several limitations, including insufficient cross-domain feature representation, inter-category ambiguity in novel classes, and inherent feature bias toward the source domain. To overcome these issues, this paper proposes a category-level collaboration knowledge mining strategy designed to comprehensively exploit both inter-class and intra-class feature relationships across domains. Specifically, a clustering-based memory bank (CMB) is initially constructed to aggregate class prototype features, class auxiliary features, and intra-class disparity features, thereby embedding rich category-level knowledge into a unified memory structure. The CMB is iteratively updated through unsupervised clustering, which facilitates the modeling of intra-category relationships and enhances its capacity for cross-domain knowledge representation. Subsequently, a base-to-novel selection metric (BNSM) is designed to identify features corresponding to novel categories within the source domain by regulating the relationships between the novel categories and each base category. The selected features are then leveraged to initialize the object detector for the classification of novel categories. Finally, an adaptive feature assignment (AFA) strategy is introduced to transfer the learned category-level knowledge to the target domain, enabling the assignment of category labels to features. The memory bank is updated asynchronously with these assigned features to mitigate source domain bias. Extensive experiments conducted on diverse domain datasets demonstrate that the proposed method consistently outperforms state-of-the-art AOOD approaches, achieving performance gains of 1.1 to 5.5 mAP. Code is available at \url{https://github.com/Jandsome/CCKM}.
\end{abstract}

\begin{IEEEkeywords}
adaptive open-set object detection, memory bank, class prototype features, class auxiliary features. 
\end{IEEEkeywords}

\section{Introduction}

\begin{figure}[!t] 
	\centering
	\includegraphics[width=\linewidth]{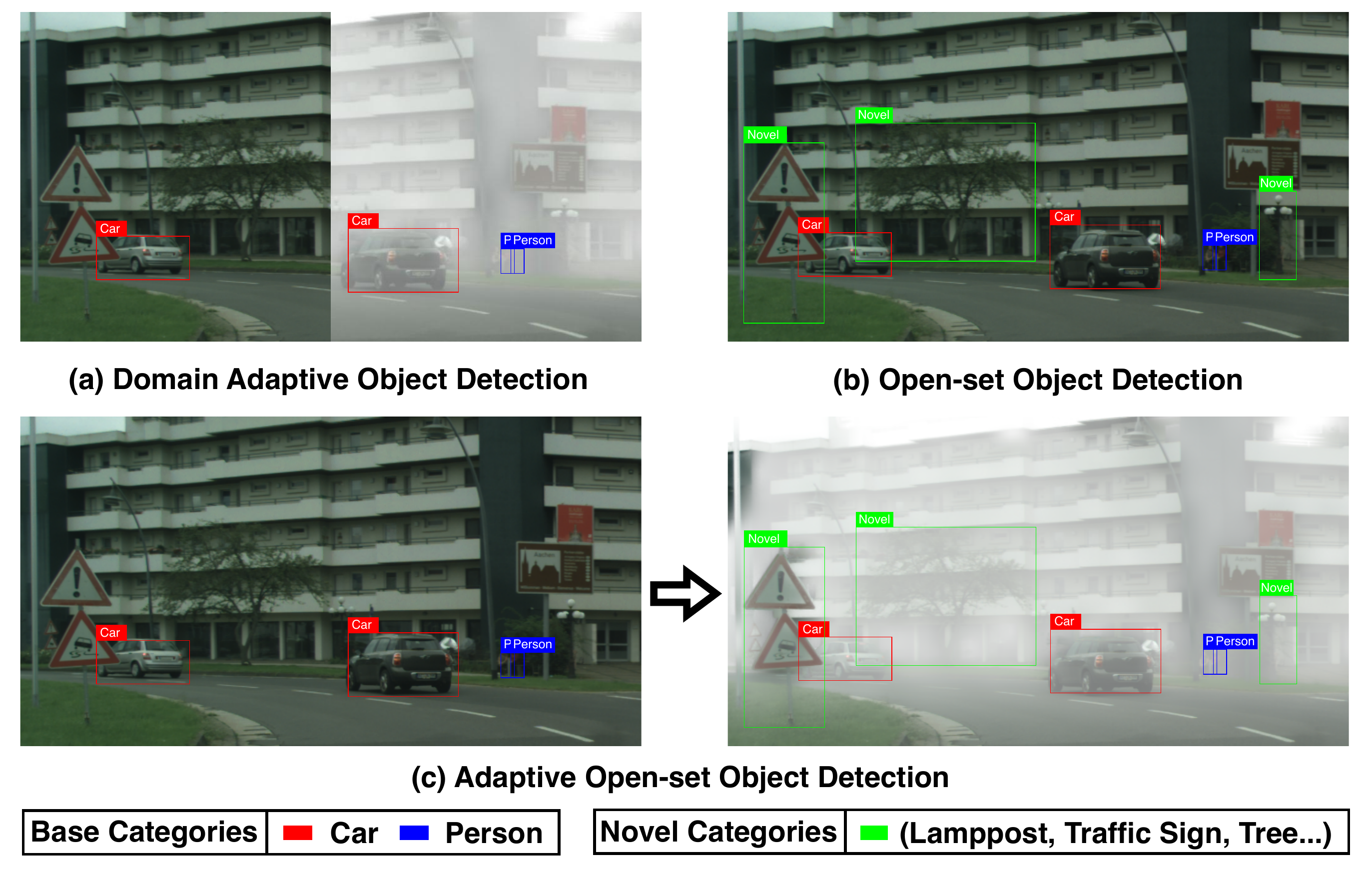}
	\caption{Illustration of (a) existing DAOD task, (b) OSOD task and (c) AOOD task. Foggy weather corresponds to the target domain, whereas clear weather corresponds to the source domain.}
\label{fig:detection branch}
\end{figure}

\IEEEPARstart
Object detection has developed rapidly and plays an essential role in various vision tasks such as image retrieval \cite{kim2024retrieval,chen2023two}, instance segmentation \cite{arnab2017pixelwise,bolya2019yolact,lee2020centermask,wan2025out}, intelligent transportation systems \cite{chen2017multi,Lin20233ddfm,su2022fsrdd}, and industrial defect detection \cite{su2022pvel}. With the continuous growth of image data, both the number of domains and object categories have increased, resulting in high manual annotation costs. Directly deploying well-trained detectors on new data domains and novel object categories leads to significant performance degradation \cite{shimodaira2000improving}. To tackle these challenges, various object detection methods have been proposed, including domain adaptive object detection (DAOD) and open-set object detection (OSOD). As depicted in Fig. \textcolor{red}{\ref{fig:detection branch}} (a) and (b), DAOD methods \cite{lin2022prototype,wang2022robust,li2022scan++,li2022sigma} are designed to train detectors exclusively on the source domain and subsequently generalize them to unseen target domains, whereas OSOD methods \cite{dhamija2020overlooked, han2022expanding, su2024toward,su2023hsic,wu2024boosting,liang2023unknown} train detectors to recognize novel object categories. As illustrated in Fig. \textcolor{red}{\ref{fig:detection branch}} (c), adaptive open-set object detection (AOOD) simultaneously performs DAOD and OSOD in an unsupervised manner.

The structured motif matching (SOMA) framework \cite{li2023novel} as a state-of-the-art AOOD method, is primarily built upon a deformable DETR \cite{zhudeformable} architecture. It integrates the training strategies employed by prototype-based DAOD methods \cite{zhang2021rpn, tian2021knowledge} and pseudo-label based OSOD methods \cite{zheng2022towards,joseph2021towards}. Specifically, SOMA first utilizes the object query features of DETR to match the ground truth in the source domain. The matched object query features are assigned to base categories, whereas the unmatched are selected and utilized for novel category identification. Subsequently, category-level knowledge of both base and novel categories is extracted from the source domain and used to select high-quality object query features in the target domain through pseudo-labeling. Finally, classification losses are calculated based on the selected object query features to optimize the detector for the target domain. However, despite the promising performance of SOMA, there exist some inherent limitations that lead to suboptimal results, which are detailed as follows.

\noindent \textbf{Limited feature representation across domains}

Current methods \cite{lin2022prototype,tanwisuth2021prototype} rely heavily on feature centroids to represent the prototype features for each category. This representation is vital for distinguishing features of the same category in the target domain. However, feature centroids become less effective when faced with significant intra-category variance in feature distributions across domains. To address this, SOMA \cite{li2023novel} uses feature centroids as prototype features and incorporates intra-category variance to capture extreme features. The combination of feature centroids and extreme features can be used to enhance discrimination. Nevertheless, the feature distributions of different categories may still exhibit similar statistical variances \cite{rousseeuw1987silhouettes,meilua2003comparing,yang2021exploiting}. Therefore, feature centroids and variance cannot be solely relied upon for effective feature discrimination. Richer category-level representations should be explicitly mined to enable more reliable feature discrimination across domains.

\noindent \textbf{Inter-category ambiguity of novel categories}

\begin{figure}[!t] 
	\centering
	\includegraphics[width=\linewidth]{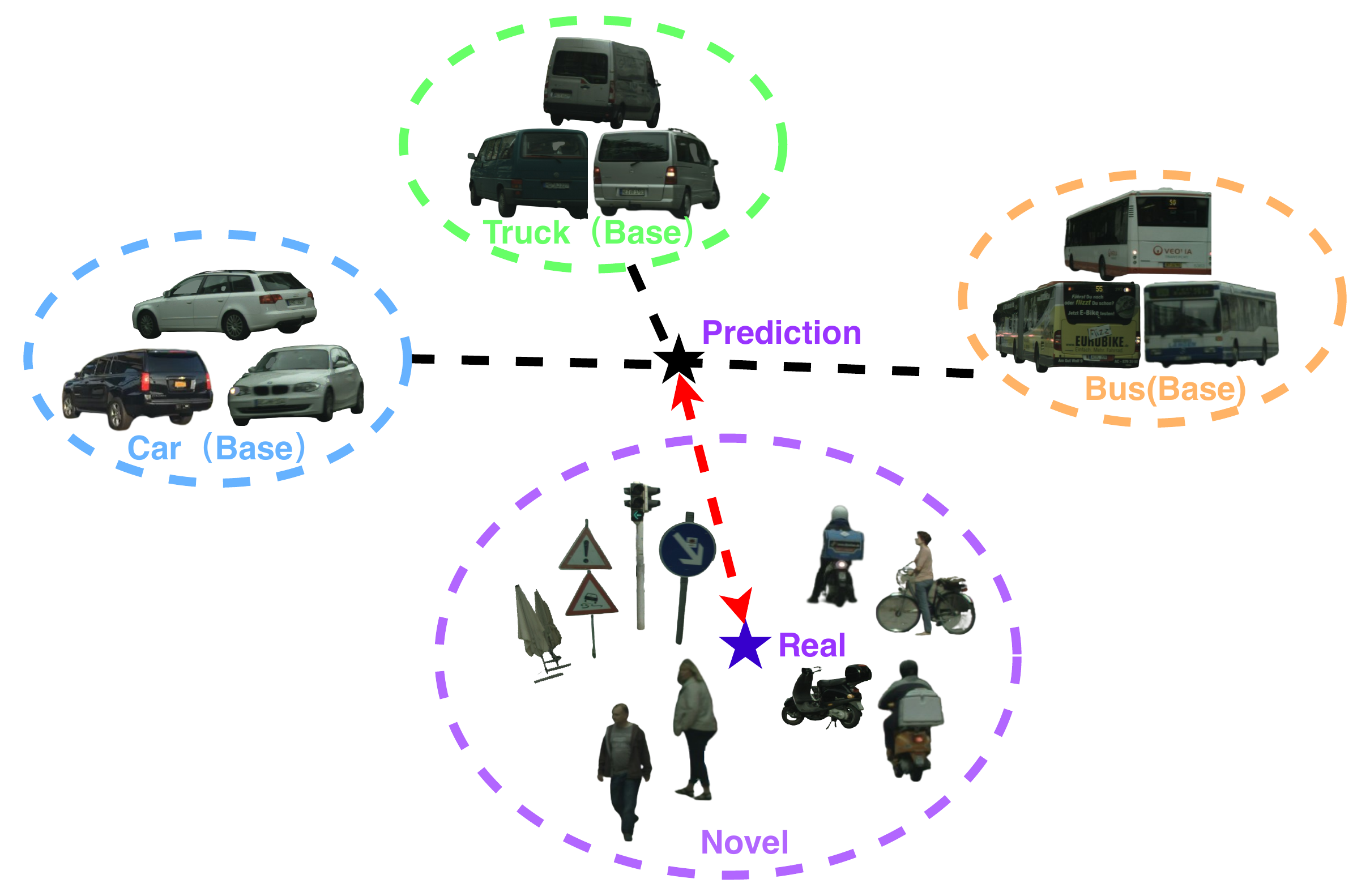} 
	\caption{Conceptual Visualization of the gap between mean features of base categories (novel category prediction) and real novel category features.}
	\label{fig:ambiguity}
\end{figure}

Several previous methods \cite{han2022expanding,liu2019separate,liang2023unknown} suggest that novel category features are closer to base category features than to the background in the feature space. Accordingly, these methods compute the mean feature representations across all base categories to represent novel categories. However, as demonstrated in Fig. \textcolor{red}{\ref{fig:ambiguity}}, mean features may not adequately represent the novel categories. To address this, SOMA \cite{li2023novel} selects object query features equidistant to the prototype features for a pair of base categories to represent novel categories. While this approach may encounter the issue illustrated in Fig.  \textcolor{red}{\ref{fig:distance}} (a), the prototype features for a pair of base categories are excessively close, leading to ambiguity in distinguishing novel categories. Hence, it is essential to establish a metric for selecting novel category features that are not too close to either the base categories or the background.

\noindent \textbf{Feature bias towards the source domain}

Due to the scarcity of high-quality pseudo-label features in the target domain, the prototype features are typically updated using features from the source domain \cite{zhang2021rpn,vs2021mega,zheng2020cross}. However, these approaches ignore the dynamic changes in feature distribution, which can result in less robust adaptation and a heavy bias towards the source domain. Hence, high-quality pseudo-label features in the target domain must be assigned to balance the update process.

To address the above limitations, category-level knowledge mining (CCKM) has been designed for AOOD. Specifically, clustering-based memory bank (CMB) incorporates class prototype features, class auxiliary features, and intra-class disparity features to construct a memory bank that stores category-level knowledge across domains. Each component of CMB is updated through unsupervised clustering, which comprehensively considers the relationships among features at the category-level. CMB enhances the representational capacity of the memory bank, serving as a bridge across domains. To mitigate ambiguity of novel categories, base-to-novel selection metric (BNSM) is employed to improve the selection of object query features in the source domain. BNSM regulates the distance between object query features of novel categories and class prototype features of base categories via dual prototype ball (ProtoBall) distance, ensuring they are neither too close nor too far apart. Consequently, BNSM contributes to improved classification performance for novel categories. Finally, to balance the feature bias,  adaptive feature assignment (AFA) assigns pseudo-labels by calculating Euclidean distance between class auxiliary features and object query features in the target domain. The object query features with pseudo-labels are then utilized in asynchronous memory bank update.  AFA ensures that the memory bank remains unbiased toward the source or the target domain.

We demonstrate the effectiveness of the proposed CCKM on various domain datasets trained without annotations. The proposed method achieves state-of-the-art performance on the BDD100k \cite{yu2020bdd100k}, Cityscapes \cite{cordts2016cityscapes}, Foggy  Cityscapes \cite{sakaridis2018semantic}, Pascal VOC \cite{everingham2015pascal} and Clipart \cite{inoue2018cross} datasets. To sum up, the key contributions of this work are as follows\\
\begin{enumerate}
	\item We identify and summarize current challenges in AOOD and propose knowledge mining via category-level collaboration knowledge mining composed of clustering-based memory bank, base-to-novel selection metric and adaptive feature assignment.
	\item Clustering-based memory bank incorporates class prototype features, class auxiliary features, and intra-class disparity features to stores category-level knowledge. It is updated through unsupervised clustering which enables the mining and transfer of category-level knowledge across domains.
	\item Base-to-novel selection metric  mitigates ambiguity in novel categories by regulating the distance between object query features of novel categories and class prototype features of base categories, thereby improving classification performance for novel categories.
	\item Adaptive feature assignment balances memory bank bias by assigning pseudo-labels and updating the memory bank asynchronously to ensure unbiased updates across both domains.\\
	\item Extensive ablation and comparison experiments are carried out on four cross-domain object detection datasets. Our method achieves state-of-the-art performance in both qualitative and quantitative comparisons across diverse challenging conditions.
\end{enumerate}

The remainder of this paper is organized as follows. Section \textcolor{red}{\ref{sec:related work}} discusses foundations of AOOD. Then the proposed CCKM, followed by CMB, BNSM, and AFA, is presented in Section \textcolor{red}{\ref{sec:proposed}}. The implementation details and experimental results are shown in Section \textcolor{red}{\ref{sec:experiment}}. Finally, Section \textcolor{red}{\ref{sec:discuss}} concludes our work and discusses the future research directions.

\begin{figure*}[!ht]
	\centering
	\includegraphics[width=\textwidth]{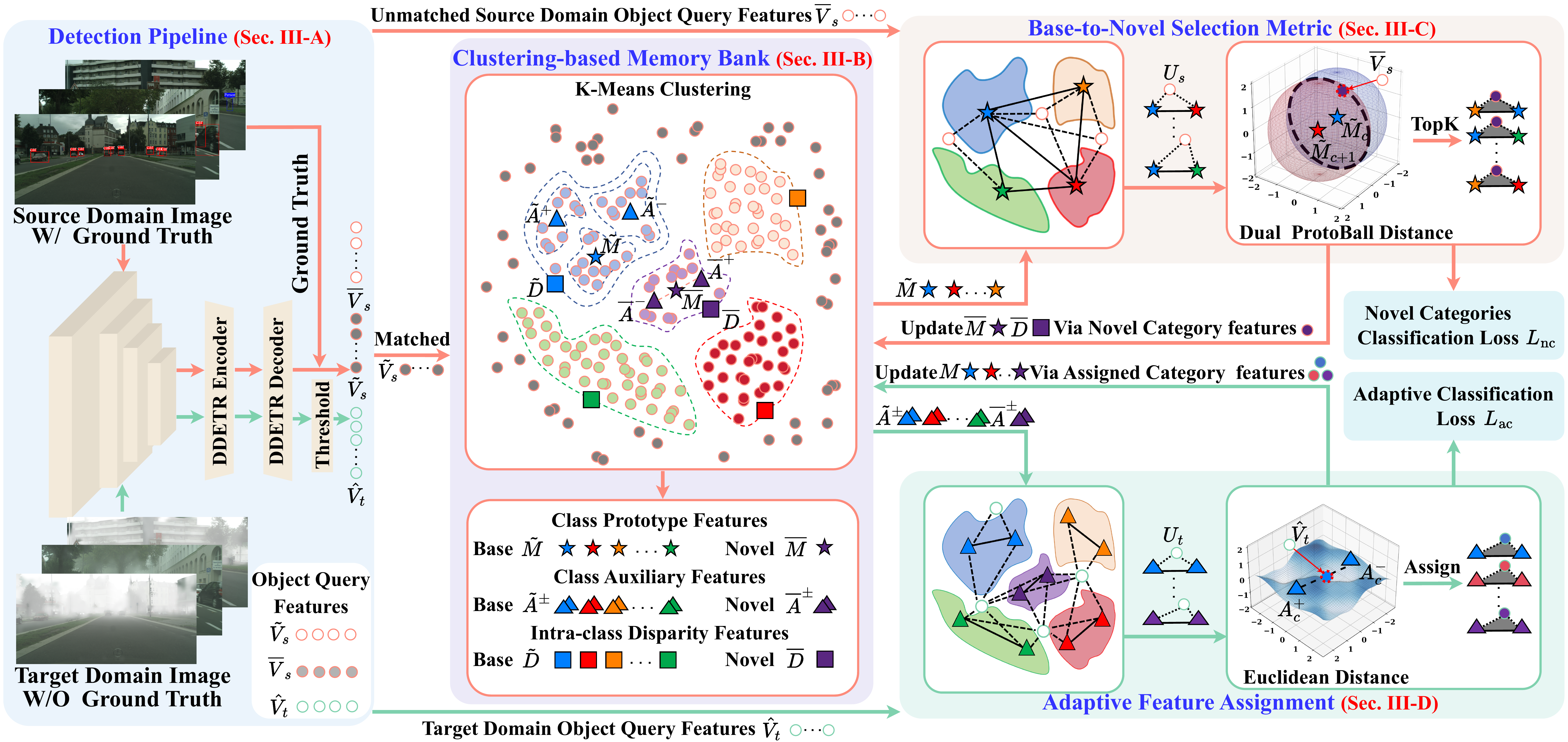}
	\caption{An overview of the proposed method. In each mini-batch, images from the source domain (with ground truth annotations) and the target domain are used as input. A shared backbone and shared detector, based on Deformable DETR (DDETR), extract image-level and object query features. Object query features from the source domain are matched with ground truth to identify features corresponding to base categories. These matched features are then used to construct a clustering-based memory bank. Class prototype features guide the base-to-novel selection metric to identify novel category features, while class auxiliary features support adaptive feature assignment by generating high-quality pseudo-labels for object query features in the target domain.}
	\label{fig:soma_pipeline}
\end{figure*}

\section{Related Work} \label{sec:related work}

\subsection{Domain Adaptive Object Detection} 
Early domain adaptive object detection (DAOD) methods such as DAF \cite{chen2018domain}, MAF \cite{he2019multi}, and SCDA \cite{zhu2019adapting} primarily rely on adversarial feature alignment, thereby limiting their capacity to model class-conditional distributions. To further improve semantic consistency, prototype-based methods \cite{zhang2021rpn, vs2021mega,zheng2020cross} introduce category-level alignment, a design that inherently constrains prototype updates to source-domain features. More recently, SIGMA \cite{li2022sigma} and SIGMA++ \cite{li2023sigma++} leverage graph matching for fine-grained cross-domain alignment, but their semantic nodes are still largely constructed from source-domain statistics. This persistent reliance on source-derived prototypes introduces inherent source bias, motivating the Adaptive Feature Assignment (AFA) to integrate target-domain features into category-level semantics.

\subsection{Open-set Object Detection}
Open-set object detection (OSOD) \cite{dhamija2020overlooked} aims to detect known objects while identifying unseen ones. FOOD \cite{su2024toward} pioneers the extension of OSOD to the few-shot setting and further enhances unknown rejection in FOODv2 \cite{su2023hsic} via HSIC-based Moving Weight Averaging. CED-FOOD \cite{wu2024boosting} further advances this line by sharpening the decision boundary with a prompt-driven mechanism. Meanwhile, UnSniffer \cite{liang2023unknown} introduces the UOD-Benchmark and a robust unknown–background separation strategy. Despite these notable advancements, most OSOD methods \cite{chen2020learning,han2022expanding,chen2021adversarial,joseph2021towards,zheng2022towards,gupta2022ow,dhamija2020overlooked} typically adopt limited category-level representations, inevitably discarding fine-grained cues. In this work, we design the Clustering-based Memory Bank (CMB) to store richer category-level representations.

\subsection{Adaptive Open-Set Object Detection}
Adaptive open-set object detection (AOOD) bridges both DAOD \cite{zhang2021rpn, tian2021knowledge} and OSOD \cite{zheng2022towards,joseph2021towards} by simultaneously handling domain shift and novel target-domain classes. While prior studies \cite{panareda2017openset, saito2018openset,  bucci2020rotation, jing2021balanced} validate cross-domain recognition of novel classes, their image-level focus provides limited insight into AOOD at the instance-level. Building on graph-motif modeling \cite{hung2017scene,benson2016higher,li2022sigma,li2023sigma++} of high-order category–object relations, SOMA \cite{li2023novel} constructs a structural metric to separate novel target-domain instances from background \cite{han2022expanding,liu2019separate,liang2023unknown}, forming the first AOOD framework. However, this metric may cause feature overlap between base and visually similar novel classes. We therefore propose the Base-to-Novel Selection Metric (BNSM) to separate novel classes from background without sacrificing base-class detection performance.

\section{The proposed Method} \label{sec:proposed}
\textbf{Task Format for AOOD: } In this section, we provide a detailed description of  AOOD task. In contrast to DAOD, AOOD relaxes the assumption that the source and target domains share the same category space. Specifically, AOOD modifies the training process of the detector to recognize shared categories across domains while enabling the classification of novel categories exclusive to the target domain \cite{li2023novel}. Let $X_{s/t}$ denote the input images in each training mini-batch data, where $s$ and $t$ refer to the source domain and target domain, respectively. The only available ground truth during training are $(Y_{s}, B_{s})$, which consist of the coordinates of the bounding boxes $B_s$ along with their corresponding category labels $Y_s \in \{1,2,\dots,C\}$ in the source domain.   Notably, no ground truth are available for the target domain.  We define $\{1,2,\dots,C\}$ as base categories. The primary objective of the proposed method is to train a detector that not only generalizes effectively on the base categories in the target domain but also uniformly classify all novel categories $\{C+2,\dots,C+C^{\prime}\}$ into a single unified novel category labeled  $C+1$.  

Fig. \textcolor{red}{\ref{fig:soma_pipeline}} shows the overview of the proposed method, which mainly consists detection pipeline, clustering-based memory bank, base-to-novel selection metric and adaptive feature assignment. In the detection pipeline, the backbone ResNet-50 first extracts image-level features from the input images across domains. Then, the image-level features and object query features are fed into the encoder-decoder module of deformable DETR (DDETR) \cite{zhudeformable} for detection. Finally, object query features are utilized to detect objects belonging to novel categories. During evaluation, only the detection pipeline is available for predicting objects from both base and novel categories using images from the target domain. We describe each subsection in detail below.
\subsection{Detection Pipeline} \label{sec:Detection pipeline}
 During the detection pipeline, the backbones and FPN \cite{lin2017feature} serve as feature extractors $\phi$, responsible for extracting image-level features $P_{s/t}$ from the source and target domains. The process is formulated as follows
\begin{eqnarray}\label{eq: feature extractor}
\begin{array}{l}
P_{s/t}= \phi(X_{s/t}),\\
\end{array}
\end{eqnarray}
where $\phi$ denotes the shared backbone (ResNet-50) that employs the same weight parameters across domains. Following the previous domain adaptive object detection method \cite{li2022sigma}, the image-level features $P_{s/t}$ are adopted to perform global adaptation. The global adaptation loss is formulated as
\begin{eqnarray}\label{eq:GA loss}
\begin{array}{l}
L_{\text{{ga}}}\hspace{-1pt}=\hspace{-1pt}-\hspace{-1pt}\left(D_{\text{{ga}}}\cdot log\left(1\hspace{-2pt}-\hspace{-2pt}\delta\hspace{-2pt}\left(P_{s}\right)\right)\hspace{-1pt}+\hspace{-1pt}\left(1\hspace{-2pt}-\hspace{-2pt}D_{\text{{ga}}} \right) \cdot log\left(\mathcal\delta\hspace{-2pt}\left(P_{t}\right)\right)\right),
\end{array}
\end{eqnarray}
where $D_{\text{{ga}}} \in \{0,1\}$ are domain labels of image-level features $P_{s/t}$. $\delta$ denotes the domain discriminators formed by binary classifiers. The global adaptation process ensures that the feature extractor can better extract domain-invariant information from image-level features.

Then, DDETR \cite{zhudeformable} utilizes object queries $V_{s/t}$ to interact with image-level features $P_{s/t}$ via encoder-decoder $\psi$. Hence, we can acquire refined object query features $V^{\prime}_{s/t} \in \mathbb{R}^{\in {N_{s/t}} \times 256}$ as instance-level features. The above process is formulated as
\begin{eqnarray}\label{eq: feature extractor}
\begin{array}{l}
V^{\prime}_{s/t}= \psi(P_{s/t}, V_{s/t}).\\
\end{array}
\end{eqnarray}

Then, the refined object queries $V^{\prime}_{s} \in \mathbb{R}^{\in {N_s}^{\prime} \times 256}$ in the source domain are fed into the detection head for regression and classification. In the meantime, Hungarian matching algorithm \cite{kuhn1955hungarian}  is employed to match the detection results with ground truth ${Y_{s},B_s}$ based on the regression and classification results as
\begin{equation}\label{eq:detection_head}
\begin{aligned}
Y_s^{\prime} &= \tau_{\text{cls}}(V^{\prime}_{s})\\
B_s^{\prime} &= \tau_{\text{reg}}(V^{\prime}_{s})\\
\tilde{V}_{s} &= \text{Hungarian}(Y_s^{\prime},  B_s^{\prime} , Y_s,B_s), \\
\end{aligned}
\end{equation}
where $\tau$ denotes the detection head. It comprises a regression head $\tau_{\text{reg}}$ that outputs bounding box predictions and a classification head $\tau_{\text{cls}}$ that produces category predictions. $Y_s^{\prime}$ and $B_s^{\prime}$ represent the category classification results and bounding box regression results, respectively. The $\text{Hungarian}$ operator refers to the Hungarian algorithm \cite{kuhn1955hungarian} for detection results assignment. After assignment, the matched assign results $Y_s^{\prime}$ and $B_s^{\prime}$ in the source domain are utilized to calculate the supervised detection loss as
\begin{equation}\label{eq:det loss}
\begin{aligned}
L_{\text{det}} &= L_{\text{reg}}(B_s^{\prime}, B_s) + L_{\text{cls}}(Y_s^{\prime}, Y_s), \\
\end{aligned}
\end{equation}
where ${L}_{\text{{reg}}}$ denotes the \emph{$GIoU$} loss  \cite{rezatofighi2019generalized} for coords localization. ${L}_{\text{{cls}}}$ denotes the focal loss \cite{lin2017focal} for object classification. 

Based on the above assignment, we can determine which object queries match the ground truth. Consequently, $\tilde{V}_{s} \in \mathbb{R}^{\in \tilde{N}_s \times 256}$ are the matched object query features for base categories, while the unmatched object query features $\overline{V}_{s} \in \mathbb{R}^{\in \overline{N}_s \times 256}$  denote the novel categories and background. The unmatched object query features $ \overline{V}_{s}=V^{\prime}_{s} \setminus \tilde{V}_{s}$ are the set difference between the refined object query features $V^{\prime}_{s}$ and the matched object queries $\tilde{V}_{s}$. In practice, $\tilde{V}_{s}$ within a mini-batch are retained, with their number dynamically reflecting the source-domain instance distribution.

The detection pipeline is designed to calculate the supervised detection loss and global adaptation loss for DDETR \cite{zhudeformable}. In the following section, the matched and unmatched object query features are utilized to identify the novel categories in the source domain. 
\subsection{Clustering-based Memory Bank} \label{sec:CBM}
Considering that the matched object query features $\tilde{V}_{s}$ are related to the base categories $\{1,2,\dots,C\}$, the unmatched object queries $\overline{V}_{s}$ correspond to the novel categories $\{C+2,\dots,C+C^{\prime}\}$ and background. We establish the clustering-based memory bank that serves as a bridge between base and novel categories for identifying object query features across domains. \textbf{Class prototype features}, denoted as $\tilde{M} \in \mathbb{R}^{C \times 256}$ for base categories and $\overline{M} \in \mathbb{R}^{1 \times 256}$ for novel categories, are designed to capture the feature centroids of each category. \textbf{Class auxiliary features}, $\tilde{A}^{\pm} \in \mathbb{R}^{2 \times C \times 256}$ and $\overline{A}^{\pm} \in \mathbb{R}^{2 \times 1 \times 256}$ for base and novel categories, capture secondary representative sub-centroids to complement class prototype features. \textbf{Intra-class disparity features}, $\tilde{D} \in \mathbb{R}^{C \times 256}$ and $\overline{D} \in \mathbb{R}^{1 \times 256}$ for base and novel categories, are constructed to encode the intra-class variability of object query features.

\begin{figure}[!t] 
	\centering
	\includegraphics[width=\linewidth]{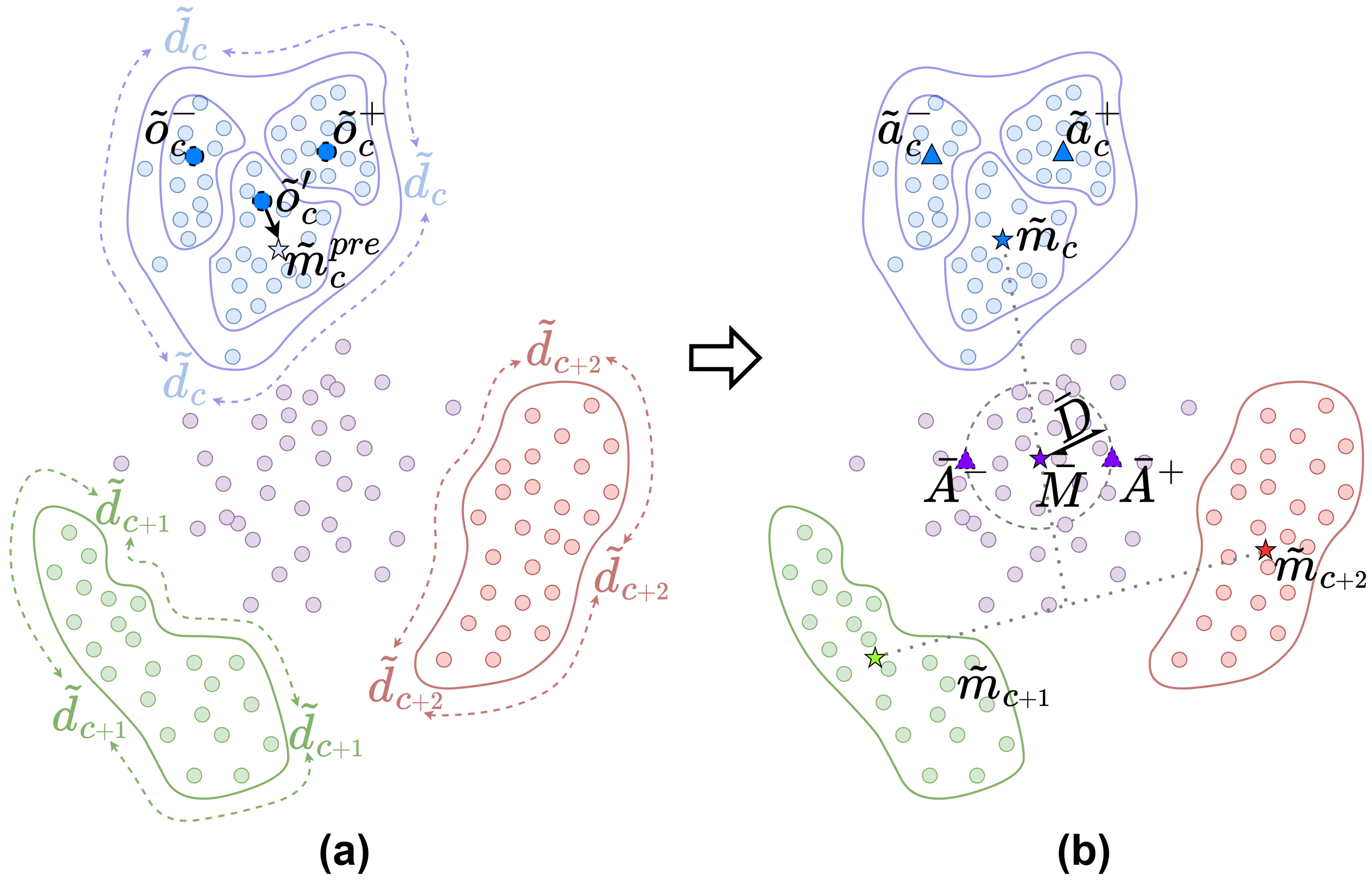} 
	\caption{Update procedure of the clustering-based memory bank (CMB). (a) For base category, prototype features and auxiliary features are updated via K-means clustering. (b) For novel category, prototype features and intra-class disparity features are derived from base class statistics.}
	\label{fig:CMB}
	\vspace{-0.3em}
\end{figure}

We establish CMB based on $\{\tilde{M},\overline{M}, \tilde{A}^{\pm}, \overline{A}^{\pm}, \tilde{D},\overline{D}\}$ to store richer category-level representations. Initially, all these features are set randomly \cite{li2022sigma, li2023sigma++, li2023novel} and updated iteratively based on object query features in each mini-batch data. \footnote{As the memory bank is continuously updated through batch-wise clustering, the overall performance is not sensitive to the specific initialization.} The details of memory bank calculation are described in Alg. \textcolor{red}{\ref{alg:base update}}. Here, $\beta$ serves as a momentum parameter \cite{joseph2021towards,he2020momentum} that balances the contribution between the historical representations and the newly aggregated features. As shown in Fig. \textcolor{red}{\ref{fig:CMB}}, we first update the $\tilde{m}_{c} \in \tilde{M}$, $\tilde{a}_{c}^{\pm} \in \tilde{A}^{\pm}$ and $\tilde{d}_{c} \in \tilde{D}$ for base category $c$. Specifically, the matched object query features $\tilde{v}_{s,c} \in \tilde{V}_{s}$ from the source domain are concatenated with the base class prototype features  $\tilde{m}_{c}$ and then perform K-\emph{means} clustering \cite{kumar2010clustering} to separate them into three clusters. The cluster $\tilde{O}_{c}^{\prime}$ containing the previous $\tilde{m}_{c}$ is selected for updating class prototype features by calculating the cosine similarity as update momentum. The mean features of the other two clusters $\{\tilde{O}_{c}^{+},\tilde{O}_{c}^{-}\}$ are directly assigned as the updated class auxiliary features $\tilde{A}^{\pm}$ for base category $c$. The intra-class disparity features $\tilde{D}$  are also updated based on standard deviation $\tilde{q}_{s,c}$. As for the novel categories, inspired by OpenDet\cite{liu2019separate} and MLFA \cite{ liu2024mlfa }, the novel class prototype features $\overline{M}$ are calculated using the mean of the base class prototype features $\tilde{M}$. The novel class auxiliary features $\overline{A}^{\pm}$ are calculated based on $\overline{M}$ and $\overline{D}$. Since CMB maintains only category-level representations and is updated with lightweight clustering, it incurs minimal computational and memory overhead, without introducing additional inference cost.

\begin{algorithm}[!t] 
	\caption{Clustering-based Memory Bank Calculation}
	\label{alg:base update}
	\begin{algorithmic}[1]
		\Require{}
		\Statex $\tilde{V}_{s}:$ Matched object query features 
		\Statex \emph{Base Categories}
		\Statex $\tilde{M}:$  Class prototype features 
		\Statex $\tilde{A}^{\pm}:$  Class auxiliary features
		\Statex $\tilde{D}:$ Intra-Class Disparity features
		\Statex \emph{Novel categories}
		\Statex $\overline{M}:$  Class prototype features 
		\Statex $\overline{A}^{\pm}:$  Class auxiliary features  
		\Statex $\overline{D}: $ Intra-Class Disparity features 
		\Statex \emph{Parameters}
		\Statex $\beta:$ Momentum Parameter $\beta=0.01$
		\Ensure{}
		\Statex The updated features for both base and novel categories include $\tilde{M}$, $\tilde{A}^{\pm}$, $\tilde{D}$, $\overline{M}$, $\overline{A}^{\pm}$ and $\overline{D}$ 
		\For{category $c=1,2,\dots,C$}
		\State \parbox[t]{\dimexpr\linewidth-1em\relax}{
			Select matched object query features $\tilde{v}_{s,c} \in \tilde{V}_{s}$, class prototype features $\tilde{m}_{c} \in \tilde{M}$, class auxiliary features $\{\tilde{a}_{c}^{+},\tilde{a}_{c}^{-}\} \in \tilde{A}^{\pm}$, and intra-class disparity features $\tilde{d}_{c} \in \tilde{D}$ for base category $c$.
		}
		\vspace{0.4em}
		\State Perform K-means clustering: 
		\Statex \hspace*{1em}$\{\tilde{O}_{c}^{\prime}, \tilde{O}_{c}^{+}, \tilde{O}_{c}^{-}\} = \operatorname{Kmeans}(\operatorname{Concat}(\tilde{v}_{s,c}, \tilde{m}_{c}), 3)$
		\State \parbox[t]{\dimexpr\linewidth-1em\relax}{
			Calculate mean features of cluster features $\tilde{O}_{c}^{\prime}$, $\tilde{O}_{c}^{\pm}$: \\
			\hspace*{3em}$\tilde{o}_{c}^{\prime} = \operatorname{Mean}(\tilde{O}_{c}^{\prime}), \quad
			\tilde{o}_{c}^{\pm} = \operatorname{Mean}(\tilde{O}_{c}^{\pm})$
		}
		\vspace{0.4em}
		\State Update class prototype features for base category $c$:
		\vspace{0.4em}
		\Statex \hspace*{1em}$\tilde{m}_{c} \leftarrow  
		\beta \cdot  
		\frac{\langle \tilde{o}_{c}^{\prime}, \tilde{m}_{c} \rangle}
		{\lVert \tilde{o}_{c}^{\prime} \rVert_{2} \cdot \lVert \tilde{m}_{c} \rVert_{2}}
		\cdot \tilde{o}_{c}^{\prime}
		+ \Bigl(1- \beta \cdot  
		\frac{\langle \tilde{o}_{c}^{\prime}, \tilde{m}_{c} \rangle}
		{\lVert \tilde{o}_{c}^{\prime} \rVert_{2} \cdot \lVert \tilde{m}_{c} \rVert_{2}}
		\Bigr)\cdot \tilde{m}_{c}$
		\State \parbox[t]{\dimexpr\linewidth-1em\relax}{
			Update class auxiliary features: \\
			\vspace{0.4em}
			\hspace*{5em}$\tilde{a}_{c}^{+} \leftarrow \tilde{o}_{c}^{+}, \quad \tilde{a}_{c}^{-} \leftarrow \tilde{o}_{c}^{-}$
		}
		\State \parbox[t]{\dimexpr\linewidth-1em\relax}{
			Calculate and update intra-class disparity: \\
			\hspace*{7em}$\tilde{q}_{s,c} = \operatorname{Std}(\tilde{v}_{s,c})$ 
		}
		\vspace{0.4em}
		\State Update intra-class disparity features for base category $c$:
		\vspace{-0.6em}
		\Statex \hspace*{0.5em}$\tilde{d}_{c} \leftarrow  
		\beta \cdot  
		\frac{\langle \tilde{d}_{c}, \tilde{q}_{s,c} \rangle}
		{\lVert \tilde{d}_{c} \rVert_{2} \cdot \lVert \tilde{q}_{s,c} \rVert_{2}}
		\cdot \tilde{q}_{s,c}
		+ \Bigl(1- \beta \cdot  
		\frac{\langle \tilde{d}_{c}, \tilde{q}_{s,c} \rangle}
		{\lVert \tilde{d}_{c} \rVert_{2} \cdot \lVert \tilde{q}_{s,c} \rVert_{2}}
		\Bigr) \cdot \tilde{d}_{c}
		$
		\EndFor
		
		\State Update class prototype features for novel categories:
		\vspace{0.4em}
		\Statex \hspace*{3.5em}$\overline{M} \leftarrow  
		\beta \cdot  
		\frac{ \langle \operatorname{Mean}(\tilde{M}), \overline{M} \rangle}
		{\lVert  \operatorname{Mean}(\tilde{M}) \rVert_{2} \cdot \lVert \overline{M} \rVert_{2}}
		\cdot  \operatorname{Mean}(\tilde{M})
		$
		\Statex \hspace*{3.5em}$\quad +
		(1- \beta \cdot  
		\frac{ \langle \operatorname{Mean}(\tilde{M}), \overline{M} \rangle}
		{\lVert  \operatorname{Mean}(\tilde{M}) \rVert_{2} \cdot \lVert \overline{M} \rVert_{2}})
		\cdot  \overline{M}$
		\vspace{0.4em}
		\State Update standard deviation features for novel categories:
		\vspace{0.2em}
		\Statex \hspace*{3.5em}$\overline{D} \leftarrow  
		\beta \cdot  
		\frac{\langle \operatorname{Mean}(\tilde{D}), \overline{D} \rangle}
		{\lVert \operatorname{Mean}(\tilde{D}) \rVert_{2} \cdot \lVert \overline{D} \rVert_{2}}
		\cdot \operatorname{Mean}(\tilde{D})$
		\Statex \hspace*{3.5em}$\quad +
		(1- \beta \cdot  
		\frac{\langle \operatorname{Mean}(\tilde{D}), \overline{D} \rangle}
		{\lVert \operatorname{Mean}(\tilde{D}) \rVert_{2} \cdot \lVert \overline{D} \rVert_{2}})
		\cdot \overline{D}$
		\vspace{0.2em}
		\State Calculate class auxiliary features for all novel categories:
		\Statex \hspace*{4.5em}$\overline{A}^{+}= \overline{M} + \overline{D}, \hspace*{1em}\overline{A}^{-}= \overline{M} - \overline{D}$
	\end{algorithmic}
\end{algorithm}
 
In Alg. \textcolor{red}{\ref{alg:base update}}, the novel class prototype features are calculated simply based on the mean features of the base class prototype features. However, it is challenging to use mean features to represent even a single novel category, let alone multiple novel categories. \footnote{Since the exact number of novel categories is unknown, we classify all novel categories into a single group.} As shown in Fig. \textcolor{red}{\ref{fig:ambiguity}}, We present an illustrative example scenario: when base categories (e.g., bus, truck, car) and novel categories (e.g., rider, pedestrian, bicycle) differ substantially, directly utilizing the mean features of the base categories may poorly represent the novel categories. Hence, a metric is formulated in the following section to restrict the relationship between the base and novel categories.

\subsection{Base-to-Novel Selection Metric} \label{sec:BNSM}
After the update, we need to identify the unmatched object query features for novel categories in the source domain. The updated category-level representations for novel categories can coarsely represent the feature distribution for all novel categories. Nevertheless, each novel category exhibits a distinct feature distribution, directly averaging base class prototype features may lead to suboptimal performance. Therefore, it is essential to train DDETR $\{\phi,\psi,\tau\}$ to identify novel categories by mining knowledge from unmatched object queries, which requires distinguishing novel-category object queries from background based on their relative positions to the base class prototype features. Based on the observation \cite{han2022expanding,liu2019separate,liang2023unknown}, unmatched object query features of novel categories tend to distribute closer to base class prototype features than background in the feature space.

\begin{figure}[!t] 
	\centering
	\includegraphics[width=\linewidth]{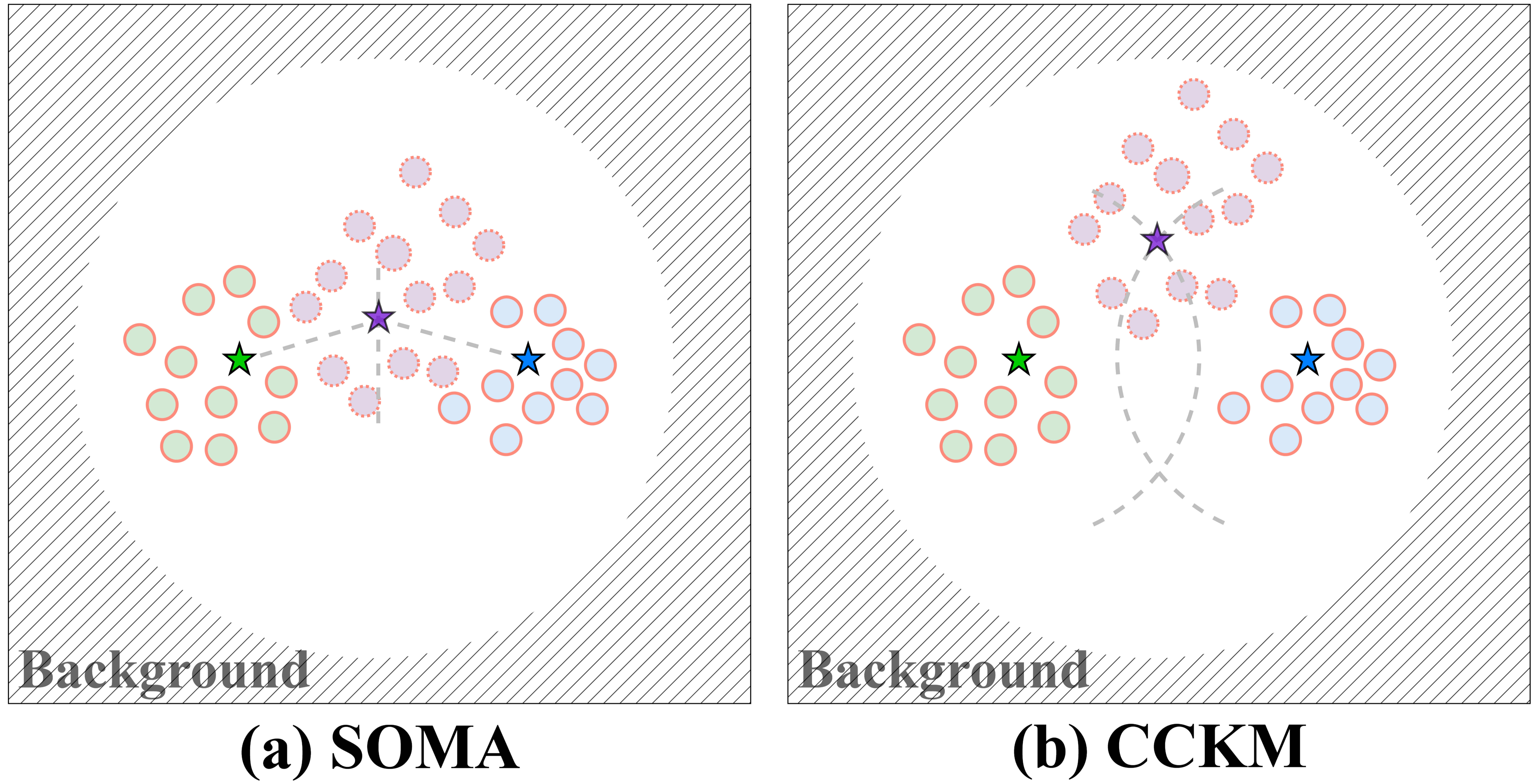}
	\vspace{-1.5em} 
	\caption{Illustration of the feature distributions for base and novel categories in (a) SOMA and (b) the proposed CCKM, respectively. The shaded regions represent background areas. The \textcolor{violet}{purple} star denotes the novel class prototype feature, while the \textcolor{green}{green} and \textcolor{blue}{blue} stars indicate two distinct base class prototype features. Compared with SOMA, CCKM exhibits reduced overlap between base and novel feature distributions.}
	\label{fig:distance}
	\vspace{-1em}
\end{figure}

Meanwhile, the feature distribution of novel categories should remain sufficiently separated from that of any base category. SOMA \cite{li2023novel} measures the relative distance between each unmatched object query features and the base class prototype features using cosine distance and NDD. Although novel categories can be distinguished from the background, their feature distributions may overlap with those of base categories, as shown in Fig. \textcolor{red}{\ref{fig:distance}}. (a). The distance between unmatched query features $\overline{v}^{\overline{n}_{s}} \in \overline{V}_{s}$ and the base class prototype features $\tilde{m}_{c}$, $\tilde{m}_{c+1}$ is measured using cosine distance. However, a small cosine distance may cause these features to be overly close to base categories in the feature space, increasing the risk of misclassification. This limitation motivates the need for a more discriminative metric. Hence, we propose a base-to-novel selection metric, as summarized in Alg. \textcolor{red}{\ref{alg:BNSM}}, to distinguish novel categories from background while reducing feature overlap with base categories. As shown in Fig. \textcolor{red}{\ref{fig:distance}}. (b) and the top right part of Fig. \textcolor{red}{\ref{fig:soma_pipeline}}, the proposed metric adopts a dual prototype ball (ProtoBall) distance, which utilizes two distinct base class prototype features as centers of balls in the feature space. Such a formulation is aligned with the principle of limiting open space risk \cite{dhamija2020overlooked,bendale2015towards,scheirer2012toward} by discouraging confident assignment of samples that lie far from known class supports, while avoiding excessive attraction to any single base class prototype feature. This dual-prototype reference design enables ProtoBall to evaluate novel queries relative to multiple base categories, alleviating bias and feature overlap with base classes. Based on the ProtoBall distance, a source domain connection matrix (SCM) is established by pairing each unmatched object query feature $\overline{v}{s}^{\overline{n}{s}} \in \overline{V}_s$, with the corresponding ProtoBall in the feature space. Each component in SCM $U_{s} \in \mathbb{R}^{C\times(C-1) \times \overline{N}_{s} }$ is formulated as follow
\begin{eqnarray}\label{eq: distance mat}
\begin{aligned}
u_{(c,c+1)}^{\overline{n}_s} &= \left|\frac{\left\|\overline{v}^{\overline{n}_{s}}-\tilde{m}_{c}\right\|_{2}-\gamma \cdot\left\|\tilde{m}_{c}-\tilde{m}_{c+1}\right\|_{2}}{\left\|\tilde{m}_{c}-\tilde{m}_{c+1}\right\|_{2}}\right| \\
& - \left|\frac{\left\|\overline{v}^{\overline{n}_{s}}-\tilde{m}_{c+1}\right\|_{2}-\gamma \cdot\left\|\tilde{m}_{c}-\tilde{m}_{c+1}\right\|_{2}}{\left\|\tilde{m}_{c}-\tilde{m}_{c+1}\right\|_{2}}\right|,
\end{aligned}
\end{eqnarray}
where $u_{(c,c+1)}^{\overline{n}_s}$ denotes the element $(c, c+1, \overline{n}_s)$ of the SCM $U_s$. It represents metric among $\tilde{m}_{c}$, $\tilde{m}_{c+1}$ and $\overline{v}^{\overline{n}_{s}}$. As illustrated in Fig. \textcolor{red}{\ref{fig:distance}} (b), the selected object query features are able to remain distinguishable from background while reducing feature overlap between base and novel categories. The scale parameter  $\gamma$ is set to 0.65. We further  investigate its optimal values in Fig. \textcolor{red}{\ref{fig:parameter analysis}} .

\begin{algorithm}[!t]
	\caption{Base-to-Novel Selection Metric}
	\label{alg:BNSM}
	\begin{algorithmic}[1]
		\Require{}
		\Statex $\overline{V}_{s}:$ Unmatched object query features from source domain
		\Statex $\tilde{M}:$ Updated base class prototype features
		\Statex \emph{Parameters}
		\Statex $K:$ Number of selected novel candidates $K=5$
		\Ensure{}
		\Statex Selected novel query features $\hat{V}_{s}$ and their indices $I$
		
		\State Initialize score vector $\overline{U}_{s} \leftarrow \mathbf{0}$
		
		\For{unmatched query features $\overline{v}^{\overline{n}_{s}} \in \overline{V}_{s}$}
	
		\For{category $c = 1,2,\dots,C$}
		\State \parbox[t]{\dimexpr\linewidth-1em\relax}{
			Select base class prototype features $\tilde{m}_{c}$, $\tilde{m}_{c+1}$ from $\tilde{M}$:
		}
		\Statex \hspace*{4.5em}
		$\tilde{m}_{c+1}
		=
		\mathop{\arg\max}\limits_{\substack{\tilde{m}_{j} \in \tilde{M}, j \neq c}}
		\left\lVert \tilde{m}_{c} - \tilde{m}_{j} \right\rVert_{2}.$	
		\State Compute ProtoBall distance $u_{(c,c+1)}^{\overline{n}_s}$
		\Statex \hspace*{12em}\textbf{$\rhd$ Defined in \textcolor{red}{Equation} (6)}
		\EndFor
		\State Compute the best-matching ProtoBall distance for $\overline{v}^{\overline{n}_s}$:
		\Statex \hspace*{6em}$\overline{U}_{s} \leftarrow \min\limits_{1 \le c < c+1 \le C} \; u_{(c,c+1)}^{\overline{n}_s}$
		\EndFor
		
		\State Select Top-$K$ novel candidates:
		\Statex \hspace*{3em}$I = \operatorname{ArgTopK}(-\overline{U}_{s}, K)$,\quad $\hat{V}_{s} = \overline{V}_{s}[I]$
		
	\end{algorithmic}
    \vspace{-0.2em}
\end{algorithm}

After obtaining the SCM $U_{s}$, we gather the smallest values for each object query features and output  $\overline{U}_{s} \in \mathbb{R}^{\tilde{N}_{s}}$. The indices of the Top-K smallest components are collected to identify high-quality object query features for novel categories in $\overline{U}_{s}$. The selection process is formulated as 
\begin{eqnarray}\label{eq: topk gathering}
\begin{array}{l}
I = \operatorname{Arg\,Topk}(-\overline{U}_{s}, K),
\end{array}
\end{eqnarray}
where $\operatorname{Arg\,Topk}$ operator is employed to collect the indices of the Top-K largest components in $-\overline{U}_{s}$, which corresponds to gathering the Top-K smallest components. The indices $I$ are utilized to select the object query features that belongs to novel categories from $\overline{V}_s$ as 
\begin{eqnarray}\label{eq: topk selection}
\begin{array}{l}
\hat{V}_s=\overline{V}_s[I],
\end{array}
\end{eqnarray}
where $\hat{V}_s$ denotes the object queries associated with novel categories. To achieve novel category recognition, the regression branch $\tau_{\text{cls}}$  of the detection head can be retrained based on the selected $\hat{V}_s$. The classification loss for novel categories is defined as follow
\begin{eqnarray}\label{eq: novel classification}
\begin{array}{l}
L_\text{nc} = -\sum \hat{Y}_s \log \tau_{\text{cls}}(\hat{V}_s),
\end{array}
\end{eqnarray}
where $L_\text{nc}$ represents the classification loss for novel categories within the source domain, while $\hat{Y}_s$ denotes the unified novel category labeled as $C+1$. In return, the classification loss contributes to the optimization of the classifier. The selected object query features $\hat{V}_s$ are representative enough for novel categories in the source domain. Hence, we utilize $\hat{V}_s$ to update the class prototype features $\overline{M}$ and intra-class disparities features $\overline{D}$ for novel categories as follows
\begin{eqnarray}\label{eq: novel classification}
	\begin{aligned}
		\overline{M} &\leftarrow  
		\beta \cdot  
		\frac{ \langle \operatorname{Mean}(\hat{V}_s), \overline{M} \rangle}
		{\lVert  \operatorname{Mean}(\hat{V}_s) \rVert_{2} \cdot \lVert \overline{M} \rVert_{2}}
		\cdot  \operatorname{Mean}(\hat{V}_s)\\
		&+ (1- \beta \cdot  
		\frac{ \langle \operatorname{Mean}(\hat{V}_s), \overline{M} \rangle}
		{\lVert  \operatorname{Mean}(\hat{V}_s) \rVert_{2} \cdot \lVert \overline{M} \rVert_{2}})
		\cdot  \overline{M},\\
		\overline{D} &\leftarrow  
		\beta \cdot  
		\frac{\langle \operatorname{Std}(\hat{V}_{s}), \overline{D} \rangle}
		{\lVert \operatorname{Std}(\hat{V}_s) \rVert_{2} \cdot \lVert \overline{D} \rVert_{2}}
		\cdot \operatorname{Std}(\hat{V}_s)\\
		&+ (1- \beta \cdot  
		\frac{\langle \operatorname{Std}(\hat{V}_s), \overline{D} \rangle}
		{\lVert \operatorname{Std}(\hat{V}_s) \rVert_{2} \cdot \lVert \overline{D} \rVert_{2}})
		\cdot  \overline{D}.
	\end{aligned}
\end{eqnarray}

The updated $\overline{M}$ and $\overline{D}$ can be utilized in the memory bank calculation in Alg. \textcolor{red}{\ref{alg:base update}} during the next iteration. High-quality $\overline{M}$ and $\overline{D}$ enhance the knowledge mining of class auxiliary features $\overline{A}^{\pm}$ for all novel categories. In the next section, the class auxiliary features $A=\{\tilde{A}^{\pm}, \overline{A}^{\pm}\}$ will be utilized to further enhance adaptation in the target domain.

\subsection{Adaptive Feature Assignment} \label{sec:AFA}
In the target domain, no ground truth labels are available for the object query features except for the pseudo labels generated by the classification branch $\tau_{\text{cls}}$. Since $\tau_{\text{cls}}$ is trained on the source domain, these pseudo labels exhibit lower confidence in the target domain due to domain shift \cite{shimodaira2000improving}. Therefore, the pseudo labels cannot be utilized as labels for training in the target domain. To address this issue, we propose an adaptive feature assignment that leverages the memory bank $\{\tilde{M},\overline{M}, \tilde{A}^{\pm}, \overline{A}^{\pm}, \tilde{D},\overline{D}\}$ to assign more accurate labels to the object query features $V_t$. In return, the assigned object query features in the target domain are  used to update the memory bank. The update process bridge the domain gap and alleviate the effects of domain shift.

According to KTNet \cite{tian2021knowledge}, features belonging to the same category should exhibit similar distributions in the feature space. Hence, the class auxiliary features $\tilde{A}^{\pm}, \overline{A}^{\pm}$  can be employed to distinguish potential foregrounds in the target domain. To select the foreground object query features from $V_{t} \in \mathbb{R}^{N_{t}\times 256}$ in the target domain, we follow the positive selection rule from DDETR \cite{zhudeformable} and set a threshold of 0.5 for foreground object query features $\hat{V}_{t} \in \mathbb{R}^{\hat{N}_{t}\times 256}$. Then, a target domain connection matrix (TCM) $U_{t} \in \mathbb{R}^{(C+1)\times \hat{N}_{t}}$ is also established between  $A^{\pm}=\{\tilde{A}^{\pm}, \overline{A}^{\pm}\}$ and $\hat{V}_{t}$. TCM $U_t$ is computed based on Euclidean distance as
\begin{eqnarray}\label{eq: adaptive classification}
\begin{aligned}
U_t = \frac{\lVert \hat{V}_{t} - \frac{{A}^{+} + {A}^{-}}{2} \rVert_{2}}{\lVert {A}^{+} - {A}^{-} \rVert_{2}}\end{aligned}.
\end{eqnarray}

Each element in TCM $U_t$ quantifies the distributional relationship between the class auxiliary features and the object query features in the feature space. Object query features are assigned to the category for which the proximity to the corresponding class auxiliary features is the closest. We determine the closest distributional relationship for each object query features by selecting the smallest values in each row of $U_t$.  Subsequently, high-quality corresponding labels $\hat{Y}_{s}$ are obtained.  It should be noted that in the target domain, the base and novel categories are computed concurrently. The adaptive classification loss is adopted for base and novel categories in the target domain as follow 
\begin{eqnarray}\label{eq: adaptive classification}
\begin{aligned}
L_\text{ac} = -\sum \hat{Y}_{t} \log \tau_{\text{cls}}(\hat{V}_t),
\end{aligned}
\end{eqnarray}
where $L_\text{ac}$ represents adaptive classification loss based on the cross entropy loss in the target domain. The classification branch $\tau_{\text{cls}}$ is trained using the loss function $L_\text{ac}$. During evaluation,  $\tau_{\text{cls}}$  determine whether features in the target domain correspond to base or novel categories. Finally, we enhance the memory bank by updating the class prototype features $M=\{\tilde{M}, \overline{M}\}$ for both base and novel categories as follow
\begin{eqnarray}\label{eq: target update}
	\begin{aligned}
		M &\leftarrow  
		\beta \cdot  
		\frac{ \langle \operatorname{Mean}(\hat{V}_{t}), M \rangle}
		{\lVert  \operatorname{Mean}(\hat{V}_{t}) \rVert_{2} \cdot \lVert M \rVert_{2}}
		\cdot  \operatorname{Mean}(\hat{V}_{t})\\
		&+ (1- \beta \cdot  
		\frac{ \langle \operatorname{Mean}(\hat{V}_{t}), M \rangle}
		{\lVert  \operatorname{Mean}(\hat{V}_{t}) \rVert_{2} \cdot \lVert M \rVert_{2}})
		\cdot  M.
	\end{aligned}
\end{eqnarray}

The updated clustering-based memory bank is utilized to enhance the category-level knowledge mining by providing richer category-level representations that are consistent across domains in Alg. \textcolor{red}{\ref{alg:base update}}. During training, it integrates these representations to capture domain-invariant characteristics of object query features across domains. This mechanism effectively improves detection performance for both base and novel categories in the target domain, while alleviating bias toward source-domain feature distributions.

\subsection{Optimization} \label{sec: loss function}
The overall objective function to train our network can be expressed as
\begin{eqnarray}\label{eq: all function}
\begin{aligned}
L = L_{\text{det}} + \lambda_{1}L_{\text{{ga}}}+\lambda_{2}L_\text{nc}+\lambda_{3}L_\text{ac}.
\end{aligned}
\end{eqnarray}

$L_{\text{det}}$  is the fully supervised detection loss in the source domain. $L_{\text{ga}}$ represents the global adaptation loss based on image-level features across domains and contributes to the extraction of domain-invariant image-level features. $L_\text{nc}$ denotes the classification loss for novel categories within the source domain and contributes to the optimization of the detection head to effectively discriminate among all novel categories. $L_\text{ac}$ describes the adaptive classification loss that enhances the cross-domain detection capabilities of the detectors. As for parameter setting, $\lambda_{1}=1e-3$, $\lambda_{2}=1e-4$ and $\lambda_{3}=1e-1$ serve as coefficients that balance the significance of the critics in the adaptation process. By such a design, the proposed method can boost the performance of AOOD.

\begin{table*}[h]
	\centering
	\renewcommand{\arraystretch}{1.2}
	\caption{Comparing with state-of-the-art validation results on Cityscapes $\rightarrow$ Foggy Cityscapes. The top 2 results are shown in \textcolor{red}{red}, \textcolor{green}{green}.}
	\resizebox{\textwidth}{!}{%
		\begin{tabular}{l|c|c c c c|c c c c|c c c c}
			\hline
			\multirow{2}{*}{\vspace{-2pt}\textbf{Method}} & \multirow{2}{*}{\vspace{-2pt}\textbf{Setting}} & \multicolumn{4}{c|}{\textbf{Num. novel categories: 3}} & \multicolumn{4}{c|}{\textbf{Num. novel categories: 4}} & \multicolumn{4}{c}{\textbf{Num. novel categories: 5}} \\ 
			\cline{3-14}
			&  & mAP $\uparrow$ & AR $\uparrow$ & WI $\downarrow$ & AOSE $\downarrow$ & mAP $\uparrow$ & AR $\uparrow$ & WI $\downarrow$ & AOSE $\downarrow$ & mAP $\uparrow$ & AR $\uparrow$ & WI $\downarrow$ & AOSE $\downarrow$\\ 
			\hline
			DDETR\cite{zhudeformable} & \multirow{6}{*}{het-sem} & 47.52 & 0.00 & 0.341 & 459 & 45.24 & 0.00 & 0.506 & 1028 & 42.38 & 0.00 & 0.659 & 1968 \\ 
			PROSER\cite{zhou2021learning}& &46.92 & 1.80 & 0.271 & 218 & 44.19 & 2.02 & 0.415 & 531 & 41.99 & 2.00 & 0.584 & 1127\\
			OpenDet\cite{han2022expanding}& &47.04 & 1.92 & 0.269 & 221 & 45.71 & 1.89 & 0.499 & 511 & 42.09 & 1.70 & 0.579 & 922\\
			OW-DETR\cite{gupta2022ow}& &43.31 & 1.84 & 0.432 & 192 & 42.52 & 2.10 & 0.619 & 451 & 39.92 & 1.98 & 0.684 & 814\\
			SOMA\cite{li2023novel}& &\textcolor{green}{50.87} & \textcolor{red}{3.78} & \textcolor{green}{0.268} & \textcolor{green}{139} & \textcolor{green}{48.06} & \textcolor{red}{4.41} & \textcolor{green}{0.412} & \textcolor{green}{340} & \textcolor{green}{45.55} & \textcolor{green}{4.08} & \textcolor{green}{0.526} & \textcolor{green}{649} \\
			\textbf{CCKM}& &\textcolor{red}{53.16} &\textcolor{green}{3.43} & \textcolor{red}{0.238} & \textcolor{red}{103} & \textcolor{red}{50.22} & \textcolor{green}{4.37} & \textcolor{red}{0.384} & \textcolor{red}{257} & \textcolor{red}{47.79} & \textcolor{red}{4.16} & \textcolor{red}{0.494} & \textcolor{red}{500}\\
			\hline
			DDETR\cite{zhudeformable} & \multirow{6}{*}{hom-sem} & 44.62 & 0.00 & 1.860 & 2937 & 43.55 & 0.00 & 2.000 & 3565 & 40.18 & 0.00 & 2.462 & 6770 \\ 
			PROSER\cite{zhou2021learning}& &43.15 & 4.59 & 1.842 & 2146 & 43.31 & 4.99 & 2.018 & 2641 & 39.99 & 5.99 & 2.563 & 4963\\
			OpenDet\cite{han2022expanding}& &45.51 & 5.28 & 1.336 & 1458 & 44.02 & 5.67 & 1.653 & 1798 & 40.87 & 6.58 & 2.303 & 3416\\
			OW-DETR\cite{gupta2022ow}& &43.22 & 3.15 & 1.355 & \textcolor{green}{1076} & 42.83 & 3.46 & 1.593 & 1320& 39.45 & 4.38 & 2.384 & 3399\\
			SOMA\cite{li2023novel}& &\textcolor{green}{48.67} & \textcolor{green}{6.96} & \textcolor{green}{1.257} & \textcolor{red}{915} & \textcolor{green}{47.02} & \textcolor{green}{7.42} & \textcolor{red}{1.527} & \textcolor{red}{1232} & \textcolor{green}{43.37} & \textcolor{green}{8.42} & \textcolor{red}{2.281} & \textcolor{red}{2886}\\
			\textbf{CCKM}& &\textcolor{red}{50.78} & \textcolor{red}{12.36} & \textcolor{red}{1.238} & 1184 & \textcolor{red}{49.71} & \textcolor{red}{12.58} & \textcolor{green}{1.558} & \textcolor{green}{1319}& \textcolor{red}{45.11} & \textcolor{red}{12.84} & \textcolor{green}{2.295} & \textcolor{green}{3356}\\
			\hline
			DDETR\cite{zhudeformable} & \multirow{6}{*}{freq-dec} & 56.99 & 0.00 & 0.579 & 1240 & 55.02 & 0.00 & 0.835 & 2136 & 53.89 & 0.00 & 0.93 & 2625 \\ 
			PROSER\cite{zhou2021learning}& &55.70 & 6.68 & 0.589 & \textcolor{red}{536} & 54.51 & 7.88 & 0.780 & 952 & 53.43 & 8.22 & 0.943 & \textcolor{red}{1072}\\
			OpenDet\cite{han2022expanding}& &57.28 & 9.35 & 0.519 & 720 & 54.89 & 10.59 & 0.781 & 1251 & 53.51 & 10.37 & 0.839 & 1470\\
			OW-DETR\cite{gupta2022ow}& &56.63 & 6.61 & 0.585 & 698 & 55.45 & 7.90 & 0.745 & \textcolor{red}{930} & 53.60 & 7.90 & 0.807 & \textcolor{green}{1105}\\
			SOMA\cite{li2023novel}& &\textcolor{green}{59.18} & \textcolor{green}{11.41} & \textcolor{red}{0.507} & \textcolor{green}{669} & \textcolor{green}{56.85} & \textcolor{green}{12.47} & \textcolor{red}{0.723} & \textcolor{green}{1140} & \textcolor{green}{55.63} & \textcolor{green}{12.36} & \textcolor{red}{0.759} & 1315\\
			\textbf{CCKM}& & \textcolor{red}{59.63} & \textcolor{red}{11.59} & \textcolor{green}{0.515} & 705 & \textcolor{red}{57.93} & \textcolor{red}{13.19} & \textcolor{green}{0.742} & 1209 & \textcolor{red}{55.74} & \textcolor{red}{13.28} & \textcolor{green}{0.802} & 1440\\
			\hline
			DDETR\cite{zhudeformable} & \multirow{6}{*}{freq-inc} & 44.72 & 0.00 & 2.862 & 2859 & 43.91 & 0.00 & 3.270 & 4907 & 41.12 & 0.00 & 3.609 & 8291 \\ 
			PROSER\cite{zhou2021learning}& &44.23 & 2.94 & 2.881 & 1090 & 42.47 & 2.98 & 2.745 & 1866 & 39.11 & 3.01 & 3.119 & 3242\\
			OpenDet\cite{han2022expanding}& &44.85 & 3.23 & 2.579 & 1700 & 42.92 & 3.30 & 2.741 & 2835 & 40.34 & 3.44 & 2.970 & 4965\\
			OW-DETR\cite{gupta2022ow}& &43.92 & \textcolor{green}{3.85} & 2.032 & 1377 & 43.01 & \textcolor{green}{3.99} & 2.219 & 1891 & 40.21 & 2.98 & 2.184 & 2293\\
			SOMA\cite{li2023novel}& &\textcolor{green}{44.30} & 3.39 & \textcolor{green}{1.398} & \textcolor{red}{394} & \textcolor{green}{44.69} & 3.55 & \textcolor{green}{1.581} & \textcolor{red}{696} & \textcolor{green}{41.16} & \textcolor{green}{3.48} & \textcolor{green}{1.800} & \textcolor{red}{1276}\\
			\textbf{CCKM}& &\textcolor{red}{46.34} & \textcolor{red}{6.10} & \textcolor{red}{1.002} & \textcolor{green}{647} & \textcolor{red}{45.14} & \textcolor{red}{6.02} & \textcolor{red}{1.167} & \textcolor{green}{1088} & \textcolor{red}{42.55} & \textcolor{red}{5.64} & \textcolor{red}{1.318} & \textcolor{green}{1896}\\
			\hline
		\end{tabular}
}
\label{tab:Foggy Cityscapes}
\vspace{-0.5em}
\end{table*}

\section{Experiments} \label{sec:experiment}
\subsection{Datasets and Evaluation Metrics}

To comprehensively evaluate the effectiveness of our approach, we conduct experiments across both street scene datasets and generic object detection datasets. 

\noindent\textbf{Street Scene Datasets}

\noindent\textbf{Cityscapes $\rightarrow$ Foggy Cityscapes.} Cityscapes \cite{cordts2016cityscapes} comprises 2,975 training images and 500 validation images of urban street scenes, with dense pixel-level annotations across 8 categories. In contrast, Foggy Cityscapes  \cite{sakaridis2018semantic} is generated by simulating fog on the Cityscapes images, presenting a challenging task for cross-domain detection. By introducing the clear-to-foggy adaptation task, we aim to evaluate the model's robustness to variations in dynamic weather conditions.

\noindent\textbf{Cityscapes$\rightarrow$ BDD100k.} BDD100K is the largest and most diverse publicly available driving dataset with 100K videos. In line with previous work \cite{wang2021exploring, wu2021vector}, we utilize the daytime subset, which includes 36,728 images for training and 5,258 images for evaluation. We assess the model's sensitivity to domain shifts induced by variations in data collection devices.

\noindent\textbf{Generic Object Detection Datasets}

\noindent\textbf{Pascal VOC $\rightarrow$ CLipart.} Pascal VOC \cite{everingham2015pascal} includes 20 object categories from real-world scenes, with 16,551 images used for training, following the mainstream \cite{chen2021i3net}. Clipart \cite{inoue2018cross} consists of 1,000 artistic style images selected from the website for training and testing \cite{saito2019strong}. The style gap between Clipart and Pascal VOC offers compelling evidence for the effectiveness of the proposed method.

To ensure a fair comparison, we evaluate detection performance on base categories in the target domain by calculating mean average precision (mAP). Specifically, AP is calculated for each class at an IoU threshold of 0.5. The mAP is then obtained by averaging these AP values across all classes. Following ORE\cite{joseph2021towards}, average recall (AR) is employed to assess the recognition performance of novel categories in the target. Higher mAP and AR values indicate the effectiveness of recognizing both base and novel categories. In addition, we employ wilderness impact (WI) to quantify the influence of unknown objects on detection performance, defined as the ratio of precision on base categories to precision on both base and novel categories. A lower WI value signifies that the presence of unknown objects has a minimal effect on the detector's precision, indicating enhanced robustness in open-set scenarios. Absolute open-set error (AOSE) quantifies the number of novel objects that are misclassified as base categories. Lower WI and AOSE values indicate that the model demonstrates robustness against a larger number of novel categories. The following section provides an in-depth description of each task.

\begin{table*}[h!]
	\centering
	\renewcommand{\arraystretch}{1.2}
	\caption{Comparing with state-of-the-art validation results on Cityscapes $\rightarrow$ BDD100k. The top 2 results are shown in \textcolor{red}{red}, \textcolor{green}{green}.}
	\resizebox{\textwidth}{!}{
		\begin{tabular}{l|c|c c c c|c c c c|c c c c}
			\hline
			\multirow{2}{*}{\vspace{-2pt}\textbf{Method}} & \multirow{2}{*}{\vspace{-2pt}\textbf{Setting}} & \multicolumn{4}{c|}{\textbf{Num. novel categories: 3}} & \multicolumn{4}{c|}{\textbf{Num. novel categories: 4}} & \multicolumn{4}{c}{\textbf{Num. novel categories: 5}} \\ 
			\cline{3-14}
			&  & mAP $\uparrow$ & AR $\uparrow$ & WI $\downarrow$ & AOSE $\downarrow$ & mAP $\uparrow$ & AR $\uparrow$ & WI $\downarrow$ & AOSE $\downarrow$ & mAP $\uparrow$ & AR $\uparrow$ & WI $\downarrow$ & AOSE $\downarrow$\\ 
			\hline
			DDETR\cite{zhudeformable} & \multirow{6}{*}{het-sem} &13.48 & 0.00 & 0.153 & 1448 & 13.49 & 0.00 & 0.164 & 1604 & 13.52 & 0.00 & 0.227 & 2378 \\ 
			PROSER\cite{zhou2021learning}& &13.32 & \textcolor{green}{1.53} & 0.148 & 910 & 13.35 & \textcolor{green}{1.48} & 0.163 & 1032 & 13.37 & \textcolor{green}{1.60} & 0.218 & 1466\\
			OpenDet\cite{han2022expanding}& &13.70 & 1.20 & 0.135 & 836 & 13.71 & 1.17 & 0.150 & 992 & 13.75 & 1.27 & 0.209 & 1244\\
			OW-DETR\cite{gupta2022ow}& &13.15 & 1.27 & 0.129 & 792 & 13.15 & 1.27 & 0.157 & 908 & 13.50 & 1.30 & 0.201 & 1168\\
			SOMA\cite{li2023novel}& & \textcolor{green}{14.11} & \textcolor{red}{1.86} & \textcolor{green}{0.127} & \textcolor{green}{614} & \textcolor{green}{14.10} & \textcolor{red}{1.90} & \textcolor{green}{0.145} & \textcolor{green}{732} & \textcolor{green}{14.13} & \textcolor{red}{2.01} & \textcolor{green}{0.197} & \textcolor{green}{1074} \\
			\textbf{CCKM}& &\textcolor{red}{14.34} &0.91 & \textcolor{red}{0.07} & \textcolor{red}{360} & \textcolor{red}{14.35} & 0.96 & \textcolor{red}{0.08} & \textcolor{red}{426} & \textcolor{red}{14.37} & 1.00 & \textcolor{red}{0.109} & \textcolor{red}{626}\\
			\hline
			DDETR\cite{zhudeformable} & \multirow{6}{*}{hom-sem} & 10.31 & 0.00 & 2.846 & 25530 & 10.32 & 0.00 & 2.873 & 26488 & 10.56 & 0.00 & 3.003 & 29812\\
			PROSER\cite{zhou2021learning}& &9.17 & 2.38 & 2.525 & 13200 & 9.19 & 2.41 & 2.458 & 13684 & 9.40 & 2.58 & 3.067 & 15962\\
			OpenDet\cite{han2022expanding}& &10.50 & \textcolor{green}{3.26} & 2.308 & 9760 & 10.54 & \textcolor{green}{3.28} & 2.327 & 10126 & 10.84 & \textcolor{green}{3.41} & 2.861 & 11776\\
			OW-DETR\cite{gupta2022ow}& &9.45 & 1.45 & 2.255 & 6236 & 9.47 & 1.46 & 2.372 & 9440 & 10.52 & 1.64 & 2.780 & 10088\\
			SOMA\cite{li2023novel}& &\textcolor{green}{11.51} & \textcolor{red}{3.97} & \textcolor{green}{2.251} & \textcolor{green}{7670} & \textcolor{green}{11.53} & \textcolor{red}{4.01} & \textcolor{green}{2.312} & \textcolor{green}{8054} & \textcolor{green}{11.83} & \textcolor{red}{4.13} & \textcolor{green}{2.611} & \textcolor{green}{9968}\\
			\textbf{CCKM}& &\textcolor{red}{11.55} & 2.26 & \textcolor{red}{1.467} & \textcolor{red}{4122} & \textcolor{red}{11.58} & 2.28 & \textcolor{red}{1.491} & \textcolor{red}{4328}& \textcolor{red}{12.06} & 2.42 & \textcolor{red}{1.861} & \textcolor{red}{5966}\\
			\hline
			DDETR\cite{zhudeformable} & \multirow{6}{*}{freq-dec} & 15.91 & 0.00 & 0.908 & 7402 & 15.88 & 0.00 & 0.952 & 8166 & 15.86 & 0.00 & 1.258 & 13044 \\ 
			PROSER\cite{zhou2021learning}& &15.98 & 12.92 & 0.949 & 4320 & 15.76 & 12.54 & 0.987 & 4886 & 12.88 & \textcolor{green}{15.57} & 1.286 & 7504\\
			OpenDet\cite{han2022expanding}& &16.01 & \textcolor{green}{14.87} & 0.948 & 4254 & 16.04 & \textcolor{green}{14.36} & 0.932 & 4942 & 16.11 & 14.69 & 1.250 & 7988\\
			OW-DETR\cite{gupta2022ow}& &15.80 & 9.68 & 0.963 & 4294 & 15.76 & 9.31 & 1.021 & 4756 & 15.81 & 9.60 & 1.379 & 7738\\
			SOMA\cite{li2023novel}& &\textcolor{green}{16.81} & \textcolor{red}{15.67} & \textcolor{green}{0.869} & \textcolor{green}{4220} & \textcolor{green}{16.55} & \textcolor{red}{15.05} & \textcolor{green}{0.915} & \textcolor{green}{4654} & \textcolor{green}{16.63} & \textcolor{red}{15.59} & \textcolor{green}{1.181} & \textcolor{green}{7230}\\
			\textbf{CCKM}& & \textcolor{red}{16.94} & 13.29 & \textcolor{red}{0.746} & \textcolor{red}{3570} & \textcolor{red}{16.94} & 12.78 & \textcolor{red}{0.784} & \textcolor{red}{3918} & \textcolor{red}{16.89} & 12.94 & \textcolor{red}{1.024} & \textcolor{red}{6152}\\
			\hline
			DDETR\cite{zhudeformable} & \multirow{6}{*}{freq-inc} & 10.02 & 0.00 & 3.054 & 22108 & 10.02 & 0.00 & 3.08 & 23060 & 10.18 & 0.00 & 3.219 & 25684 \\ 
			PROSER\cite{zhou2021learning}& &9.02 & 1.71 & 3.995 & 24118 & 8.95 & 1.72 & 4.019 & 25366 & 9.80 & 1.77 & 4.202 & 28170\\
			OpenDet\cite{han2022expanding}& &10.47 & 1.68 & 3.228 & 13578 & 10.30 & 1.70 & 3.282 & 14210 & 10.46 & 1.73 & 3.393 & 15928\\
			OW-DETR\cite{gupta2022ow}& & 8.11 & 1.75 & 2.785 & 9602 & 8.12 & 1.75 & 2.787 & 9960 & 8.34 & 1.76 & 2.867 & 11034\\
			SOMA\cite{li2023novel} & & \textcolor{green}{11.17} & \textcolor{red}{4.56} & \textcolor{red}{2.556} & \textcolor{green}{7420} & \textcolor{green}{11.08} & \textcolor{red}{4.56} & \textcolor{red}{2.577} & \textcolor{green}{7762} & \textcolor{green}{11.71} & \textcolor{red}{4.53} & \textcolor{green}{2.713} & \textcolor{green}{8844} \\
			\textbf{CCKM}& &\textcolor{red}{11.59} & \textcolor{green}{2.81} & \textcolor{green}{2.584} & \textcolor{red}{2640} & \textcolor{red}{11.51} & \textcolor{green}{3.17} & \textcolor{green}{2.653} & \textcolor{red}{2808} & \textcolor{red}{11.75} & \textcolor{green}{2.60} & \textcolor{red}{2.670} & \textcolor{red}{3286}\\
			\hline
		\end{tabular}
	}
	\label{tab:BDD100k}
	\vspace{-0.5em}
\end{table*}

\subsection{Implementation Details}
Following prior works, input images are uniformly resized to the same scale used in previous works \cite{zhou2021learning, han2022expanding, gupta2022ow, li2023novel}, while maintaining their original aspect ratios. Further implementation details are presented in the following section.

\textbf{Architecture:} The detector is implemented using Deformable DETR \cite{zhudeformable} with a ResNet-50 \cite{he2016deep} backbone. To prevent novel-class leakage from ImageNet \cite{deng2009imagenet}, as noted in \cite{gupta2022ow}, the backbone is implemented with weights pre-trained by DINO \cite{zhangdino} on the Objects365 dataset \cite{shao2019objects365}.

\textbf{Hyper-parameters:} The training phase is implemented on two NVIDIA V100 GPUs, employing the AdamW optimizer \cite{loshchilovdecoupled} with a learning rate of 0.0002, a batch size of 4, and a weight decay of 0.0005. All other hyperparameters are configured according to the default settings used in previous studies \cite{gupta2022ow, li2023novel}.

\subsection{State-of-the-art Comparison}
In this subsection, we conduct extensive experiments to compare CCKM with current SOTA methods. Following the previous works, all experimental settings remain the same as the baseline \cite{li2023novel}.

\subsubsection{Cityscapes $\rightarrow$ Foggy Cityscapes}
Table \ref{tab:Foggy Cityscapes} presents the quantitative comparison of the SOTA open-set object detection methods on the Cityscapes $\rightarrow$ Foggy Cityscapes task. Each setting varies along semantic category relationship (het-sem vs. hom-sem) or object frequency (freq-dec vs. freq-inc), while the number of novel categories ranges from 3 to 5. 

Under the heterogeneous semantics (het-sem) setting, the proposed method consistently achieves the best performance across all metrics and novel category settings. With 3 novel categories, it attains the highest base category detection performance (53.16 mAP), while maintaining a competitive classification accuracy (3.43 AR) and the lowest WI (0.238) and AOSE (103). As the number of novel categories increases to 5, the proposed method retains leading scores (47.79 mAP, 4.16 AR, WI = 0.494, AOSE = 500) and demonstrates superior scalability, outperforming SOMA \cite{li2023novel} and OpenDet \cite{han2022expanding}.

In the homogeneous semantics (hom-sem) scenario, strong semantic overlap between base and novel categories degrades base-class detection while increasing novel-class recall. Compared with SOMA, CCKM achieves higher mAP (50.78) and lower WI (1.238) in this challenging setting by explicitly reducing base–novel feature confusion through BNSM. Meanwhile, by facilitating clearer separation between novel instances and background, CCKM further improves AR (12.36). This suggests that under inevitably base–novel semantic overlap, our method aims to reduce confusion while encouraging the separation of novel instances from the background.

The frequency decrease  (freq-dec) setting simulates a long-tailed distribution where novel categories are less frequent. This imbalance is particularly challenging for novel class detection. CCKM shows the SOTA results across all configurations. For example, with 4 novel categories, it achieves a strong WI (0.742) and maintains the best AR (13.19), demonstrating resilience against data imbalance. Its performance is closely aligned with SOMA, yet consistently superior in mAP and AR, reinforcing the detection ability to generalize to rare novel classes without sacrificing base class performance.

The frequency increase (freq-inc) scenario, more frequent novel categories intensify base–novel confusion, leading to reduced mAP. Nevertheless, CCKM again surpasses all baselines, with a substantial improvement in AR (e.g., 6.10 with 3 novel categories) and the lowest WI (1.002). As novel categories become more frequent, increased intra-class complexity causes more background regions to be misclassified as novel, leading to a higher AOSE. Despite this, CCKM maintains a favorable balance between precision, recall, and open-set error.

Across all experimental settings and increasing numbers of novel categories, the proposed method achieves consistently superior performance in base category precision (mAP), novel category recall (AR), and robustness to open-set noise (low WI and AOSE). The results clearly demonstrate its capacity to adapt across semantically category diverse and frequency-imbalanced conditions, confirming its effectiveness for scalable and robust detection performance.

\begin{table}[!t]
	\renewcommand{\arraystretch}{1.25}
	\setlength\tabcolsep{3pt}
	\caption{Comparing with state-of-the-art validation results on Pascal VOC $\rightarrow$ Clipart.(num. indicates the number of novel classes.) The top 2 results are shown in \textcolor{red}{red}, \textcolor{green}{green}.}
	\begin{tabular*}{\columnwidth}{@{\hspace{2\tabcolsep}}l@{\hspace{4\tabcolsep}}|c|c@{\hspace{6\tabcolsep}}c@{\hspace{6\tabcolsep}}c@{\hspace{6\tabcolsep}}c}
		\hline
		\textbf{Method} & \textbf{Num.} & mAP $\uparrow$ & AR $\uparrow$ & WI $\downarrow$ & AOSE $\downarrow$ \\
		\hline
		DDETR\cite{zhudeformable} & \multirow{6}{*}{\textbf{6}} & 19.78 & 0.00 & 8.95 & 6347  \\ 
		PROSER\cite{zhou2021learning}& &18.23 & 32.37 & 9.87 & 5853\\
		OpenDet\cite{han2022expanding}& &20.57 & 41.15 & 8.93 & 4295 \\
		OW-DETR\cite{gupta2022ow}& &20.31 & 35.48 & 10.26 & 5184 \\
		SOMA\cite{li2023novel}& &\textcolor{green}{21.70} & \textcolor{red}{43.15} & \textcolor{green}{7.32} & \textcolor{green}{4278} \\
		\textbf{CCKM}& &\textcolor{red}{23.70} &\textcolor{green}{36.72} & \textcolor{red}{6.77} & \textcolor{red}{3496}\\
		\hline
		DDETR\cite{zhudeformable} & \multirow{6}{*}{\textbf{8}} & 19.31 & 0.00 & 9.58 & 7402  \\ 
		PROSER\cite{zhou2021learning}& &18.37 & 33.07 & 10.40 & 6636\\
		OpenDet\cite{han2022expanding}& &20.84 & 41.58 & 9.53 & \textcolor{green}{4919} \\
		OW-DETR\cite{gupta2022ow}& &21.01 & 36.53 & 10.52 & 5981 \\
		SOMA\cite{li2023novel}& &\textcolor{green}{21.69} & \textcolor{red}{43.40} & \textcolor{green}{8.24} & 5016\\
		\textbf{CCKM}& &\textcolor{red}{23.36} &\textcolor{green}{37.93} & \textcolor{red}{7.85} & \textcolor{red}{4160}\\
		\hline
		DDETR\cite{zhudeformable} & \multirow{6}{*}{\textbf{10}} & 19.12 & 0.00 & 10.06 & 9198  \\ 
		PROSER\cite{zhou2021learning}& &16.80 & 33.74 & 11.06 & 8065\\
		OpenDet\cite{han2022expanding}& &18.87 & 41.50 & 10.24 & 6103 \\
		OW-DETR\cite{gupta2022ow}& &18.42 & 36.50 & 11.06 & 7018 \\
		SOMA\cite{li2023novel}& &\textcolor{green}{20.09} & \textcolor{red}{43.73} & \textcolor{green}{8.88} & \textcolor{green}{6092} \\
		\textbf{CCKM}& &\textcolor{red}{21.99} &\textcolor{green}{38.79} & \textcolor{red}{8.11} & \textcolor{red}{5018}\\
		\hline
	\end{tabular*}
	\label{tab:Clipart}
	\vspace{-0.5em}
\end{table}

\subsubsection{Cityscapes $\rightarrow$ BDD100k} 
For the Cityscapes to BDD100k task, we adhere to the same experimental settings as those used in the Cityscapes to Foggy Cityscapes task, with the results presented in Table \ref{tab:BDD100k} .

Under the het-sem setting, CCKM sets new SOTA results across all metrics and novel category counts. It achieves the highest mAP in every case (e.g., 14.34 mAP with 3 novel categories), indicating strong detection capability on base classes. Additionally, the proposed method obtains the lowest WI (0.07) and lowest AOSE (360), indicating exceptional robustness to unknown categories. While SOMA attains higher AR, CCKM's superior precision (mAP) and drastically reduced open-set errors signify a better overall balance.

Hom-sem settings are challenging due to strong semantic overlap between base and novel classes, which leads to higher WI and AOSE for most methods. While SOMA attains higher AR by more loosely accepting novel instances, this also increases interference with base classes and background. In contrast, by integrating target-domain information through AFA and CMB, the proposed method learns more concentrated category semantics, resulting in slightly lower AR but substantially reduced WI (1.467) and AOSE (4122), and thus stronger open-set reliability.

In the freq-dec setting, which simulates the long-tail distribution, the proposed method again achieves the highest mAP (16.94) and the lowest WI and AOSE across all settings. While SOMA slightly surpasses in AR (15.67), the proposed method exhibits more consistent and robust performance. Notably, WI is reduced to 0.746, and AOSE drops to 3570, underscoring its effectiveness in handling infrequent novel instances while maintaining base class precision.

In the freq-inc setting, frequent novel occurrences intensify novel–background ambiguity, leading prior methods to misclassify background as novel. In contrast, the proposed method adopts a conservative, target-aligned detection strategy that substantially reduces false novel detections. Although AR slightly decreases (to 2.81), this is accompanied by a consistent mAP improvement (11.59) and a large reduction in open-set errors (2640 vs. 7420 for SOMA), demonstrating strong open-set robustness under frequent novel appearance.

The proposed method consistently ranks first in mAP, WI, and AOSE, while offering competitive AR. This indicates a clear advantage in base class precision, open-set robustness, and false positive suppression. The results affirm the scalability and effectiveness of the proposed model in diverse and challenging open-set scenarios, particularly under high semantic overlap and class frequency shifts.

\begin{figure}[!t] 
	\centering
	\includegraphics[width=\linewidth]{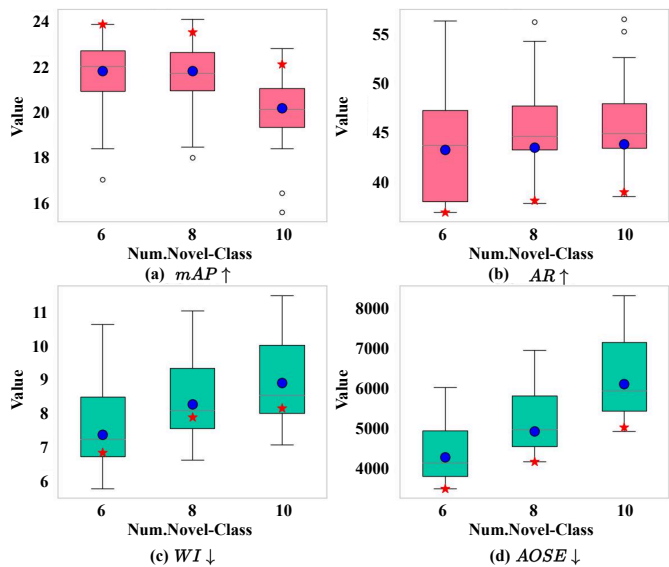} 
	\caption{The boxplot for Pascal VOC $\rightarrow$ CLipart, where the blue dots \textcolor{blue}{$\bullet$} and the red stars \textcolor{red}{$\star$} indicate the performance of the baseline and CCKM, respectively.}
	\label{fig:P_C}
	\vspace{-0.5em}
\end{figure}

\subsubsection{PascalVOC $\rightarrow$ CLipart}
As shown in Table \ref{tab:Clipart} , we conduct experiments on the Pascal VOC to Clipart task. CCKM demonstrates consistent superiority in mAP, WI, and AOSE across all settings, indicating robust detection with minimal false novel category objects. SOMA consistently ranks first in AR, showing its strength in classification, but tends to underperform in handling open-set errors. As the number of novel classes increases, WI and AOSE increase across all methods. The proposed method scales better in retaining performance, suggesting improved AOOD performance.  The performance of each metric is further illustrated in the box plot presented in Fig. \ref{fig:P_C}. We present several better results for CCKM, the majority of which exceed those of SOMA. Based on the observations from the box plot, the proposed method demonstrates superior average performance across all four metrics.  Based on these results, it is evident that CCKM exhibits excellent performance in detecting novel classes, especially in scenarios with a higher number of novel classes, while maintaining the integrity of base-class object detection in the target domain.

\begin{table}[!t]
	\renewcommand{\arraystretch}{1.25}
	\setlength\tabcolsep{3pt}
	\caption{Ablation study on Cityscapes $\rightarrow$ Foggy Cityscapes under het-sem setting (5 novel classes). The best results are highlighted in bold.}
	\begin{tabular*}{\columnwidth}{c@{\hspace{4\tabcolsep}}c@{\hspace{4\tabcolsep}}c@{\hspace{2\tabcolsep}}| c @{\hspace{7\tabcolsep}}c  @{\hspace{7\tabcolsep}}c @{\hspace{7\tabcolsep}}c}
		\hline
		\textbf{BNSM} &\textbf{CMB}  & \textbf{AFA} & mAP $\uparrow$ & AR $\uparrow$ & WI $\downarrow$ & AOSE $\downarrow$ \\
		\hline
		\multicolumn{3}{c|}{\textbf{Baseline (SOMA)}} & 45.55 & 4.08 & 0.526 & 649\\
		\hline
		\ding{52} &  &  & 46.92 & 3.15 & 0.511 & 813 \\ 
		& \ding{52} & &  45.61 & \textbf{4.34} & 0.524 & 579\\
		& & \ding{52} &  46.47 & 3.35 & 0.641 & 634\\
		\ding{52}& \ding{52} & &47.15 & 4.26 & 0.497 & 730\\
		& \ding{52}&\ding{52} &  46.38 & 3.78 & 0.601 & 525\\
		\ding{52}& &\ding{52} &   47.56& 2.55 & 0.498 & 702\\
		\ding{52}& \ding{52}&\ding{52}& \textbf{47.79} & 4.16 & \textbf{0.494} & \textbf{500}\\
		\hline
	\end{tabular*}
	\label{tab:ablation1}
	\vspace{-0.3em}
\end{table}

\begin{table}[!t]
	\setlength\tabcolsep{3pt}
	\renewcommand{\arraystretch}{1.25}
	\caption{Comparison with connection matrix (SCM : $U_{s}$ and TCM : $U_{t}$) on Cityscapes $\rightarrow$ Foggy Cityscapes het-sem setting (5 novel classes). The best results are highlighted in bold.}
	
	\begin{tabular*}{\columnwidth}{@{\hspace{4\tabcolsep}}c  @{\hspace{5\tabcolsep}}c @{\hspace{4\tabcolsep}}| c  @{\hspace{9\tabcolsep}}c  @{\hspace{9\tabcolsep}}c @{\hspace{9\tabcolsep}} c  }
		\hline
		\textbf{SCM} & \textbf{TCM} & mAP $\uparrow$ & AR $\uparrow$ & WI $\downarrow$ & AOSE $\downarrow$ \\
		\hline
		\ding{52} &   & 46.65 & 2.55 & 0.574 & 974 \\ 
		& \ding{52}  &  46.36 & \textbf{3.31} & 0.568 & 824 \\
		\ding{52}& \ding{52}  &\textbf{46.92} & 3.15 & \textbf{0.511} & \textbf{813}\\
		\hline
	\end{tabular*}
	\label{tab:ablation2}
\end{table}

\begin{table}[!t]
	\centering
	\setlength{\tabcolsep}{6pt}
	\renewcommand{\arraystretch}{1.25}
	\caption{Ablation Study on Prototype Modeling Strategies on Cityscapes $\rightarrow$ Foggy Cityscapes under het-sem setting (5 novel classes). Cosine and ProtoBall denote cosine distance and ProtoBall distance, respectively. The best results are highlighted in bold.}
	\setlength{\tabcolsep}{6pt}
	\begin{tabular*}{\linewidth}{l|c@{\hspace{6\tabcolsep}}c@{\hspace{6\tabcolsep}}c@{\hspace{6\tabcolsep}}c}
		\hline
		Constraint & mAP $\uparrow$ & AR $\uparrow$ & WI $\downarrow$ & AOSE $\downarrow$ \\
		\hline
		Cosine & 44.83 & 3.09 & 0.544 & 834 \\
        ProtoBall & \textbf{46.92} & \textbf{3.15} & \textbf{0.511} & \textbf{813} \\
		\hline
	\end{tabular*}
	\label{tab:protoball}
\end{table}

\begin{table}[!t]
	\centering
	\setlength{\tabcolsep}{6pt}
	\renewcommand{\arraystretch}{1.25}
	\caption{Sensitivity analysis of the momentum parameter $\beta$ on Cityscapes $\rightarrow$ Foggy Cityscapes under het-sem setting (5 novel classes). The best results are highlighted in bold.}
	\setlength{\tabcolsep}{6pt}
	\begin{tabular*}{\linewidth}{@{\hspace{2\tabcolsep}}c|c@{\hspace{6\tabcolsep}}c@{\hspace{6\tabcolsep}}c@{\hspace{6\tabcolsep}}c}
		\hline
		$\beta$ & mAP $\uparrow$ & AR $\uparrow$ & WI $\downarrow$ & AOSE $\downarrow$ \\
		\hline
		$1e-4$ & 45.42 & 3.26 & \textbf{0.485} & \textbf{492} \\
		$1e-3$ & 46.62 & 3.44 & 0.510 & 539 \\
		$1e-2$ & \textbf{47.79} & \textbf{4.16} & 0.494 & 500 \\
		$5e-2$ & 47.71 & 4.09 & 0.498 & 512 \\
		\hline
	\end{tabular*}
	\label{tab:beta_sensitivity}
\end{table}

\subsection{Ablation Study} \label{sec:ab exp}
In this subsection, we conduct comprehensive ablation experiments to thoroughly analyze the effect of each proposed component.

\begin{table}[!t]
		\centering
		\setlength{\tabcolsep}{6pt}
		\renewcommand{\arraystretch}{1.25}
		\caption{Sensitivity analysis of the hyperparameter $K$ on Cityscapes $\rightarrow$ Foggy Cityscapes under het-sem setting (5 novel classes). The best results are highlighted in bold.}
		\setlength{\tabcolsep}{6pt}
		\begin{tabular*}{\linewidth}{@{\hspace{2\tabcolsep}}c@{\hspace{2\tabcolsep}}|c@{\hspace{7\tabcolsep}}c@{\hspace{7\tabcolsep}}c@{\hspace{7\tabcolsep}}c}
			\hline
			$K$ & mAP $\uparrow$ & AR $\uparrow$ & WI $\downarrow$ & AOSE $\downarrow$ \\
			\hline
			$3$ & \textbf{48.13} & 3.91 & 0.560 & 631 \\
			$5$ & 47.79 & \textbf{4.16} & \textbf{0.494} & \textbf{500} \\
			$7$ & 47.54 & 4.12 & 0.538 & 528 \\
			\hline
		\end{tabular*}
		\label{tab:k_sensitivity}
\end{table}

\begin{figure}[!t] 
	\centering
	\includegraphics[width=\linewidth]{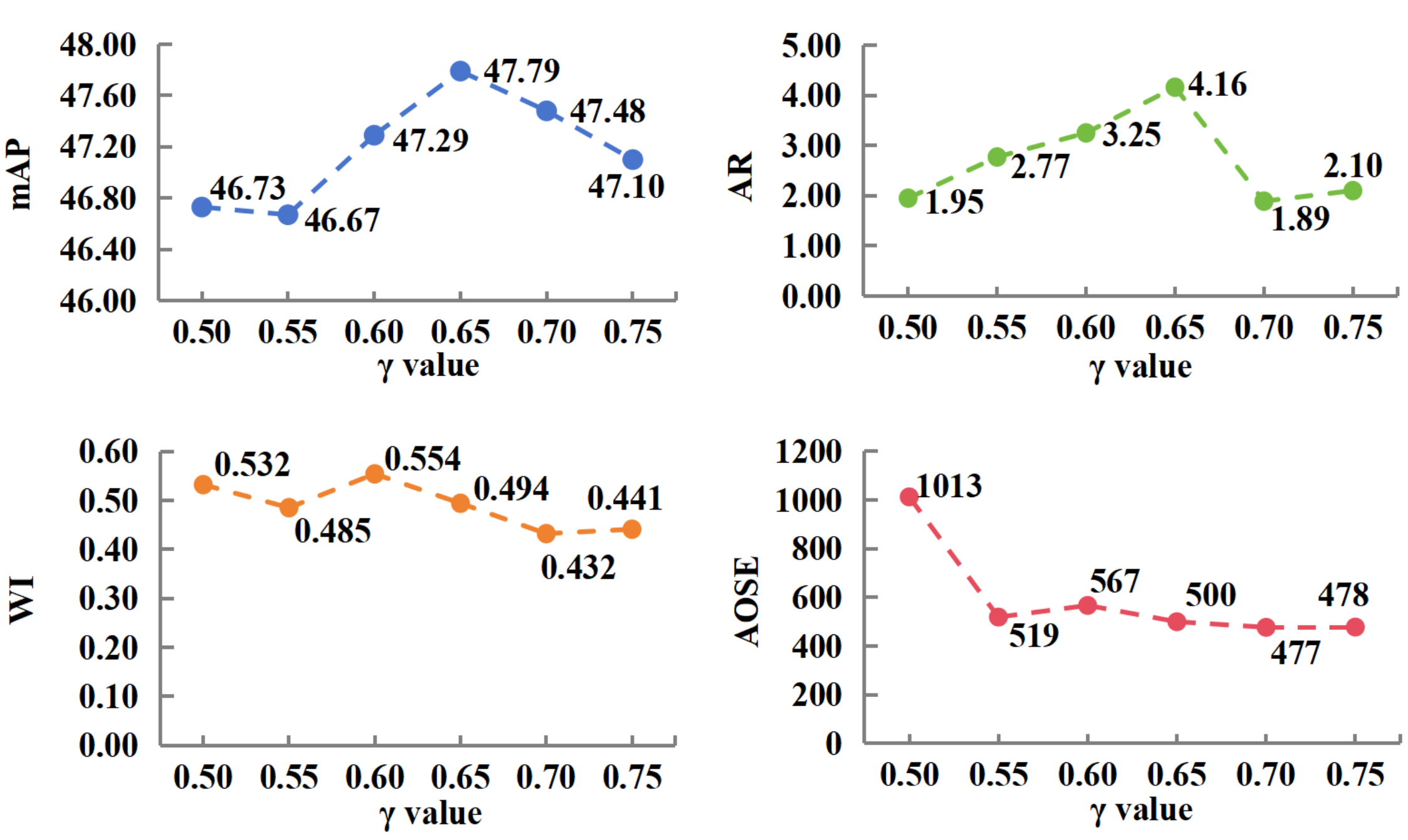}
	\caption{Sensitivity analysis of the hyperparameter $\gamma$  for Cityscapes $\rightarrow$ Foggy Cityscapes (het-sem). The plot displays the variation curves of $mAP$(top left), $AR$(top right), $WI$ (bottom left), and $AOSE$ (bottom right) as the $\gamma$ value changes.}
	\label{fig:parameter analysis}
\end{figure}

\begin{figure}[!t] 
	\centering
	\includegraphics[width=\linewidth]{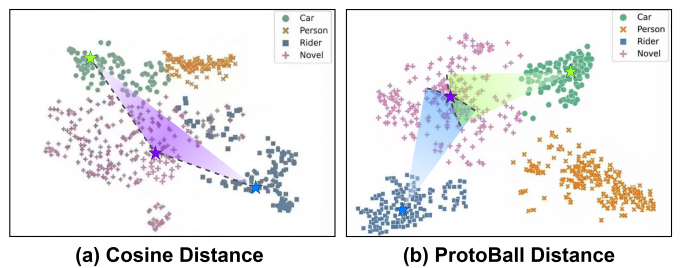} 
	\caption{t-SNE visualization of object query features on Cityscapes → Foggy Cityscapes under the freq-dec setting with 3 novel classes. (a) cosine distance; (b) ProtoBall distance for measuring unmatched object queries relative to base class prototype features.}
	\label{fig:tsne}
	\vspace{-0.5em}
\end{figure}

\subsubsection{Component-Wise Analysis}
To validate the proposed method, we conduct an ablation study on Cityscapes → Foggy Cityscapes under the het-sem setting with five novel classes in Table \ref{tab:ablation1} , using SOMA as the baseline.
Adding BNSM alone improves mAP to 46.92 but reduces AR to 3.15, as it alleviates base–novel feature confusion while discarding some novel instances that are not sufficiently distinguishable from the background. Enabling CMB alone yields consistent improvements on both base and novel categories. AR increases to 4.34 while maintaining comparable mAP (45.61), suggesting that CMB provides richer category-level representations that enhance novel instance recall without sacrificing base class reliability. Incorporating AFA alone results in a mAP of 46.47, accompanied by a decrease in AR to 3.35. This effect is attributed to AFA mitigating source-domain bias by incorporating target-domain features into memory bank updates, which stabilizes base-class predictions while excluding novel instances that fail to align with the more concentrated, target-aligned semantics. For component combinations, BNSM + CMB improves mAP to 47.15 and restores AR to 4.26, highlighting that richer category-level representations can compensate for the recall reduction introduced by BNSM. CMB + AFA achieves a lower AOSE (525) with competitive mAP (46.38) and AR (3.78), indicating improved open-set reliability. In contrast, BNSM + AFA attains strong base-class performance (47.56 mAP) but significantly reduces AR to 2.55, as stricter constraints further limit novel instance acceptance. When all components are integrated, the model achieves the best overall performance, demonstrating their complementary effects.

\begin{figure*}[!t]
	\centering
	\includegraphics[width=\textwidth]{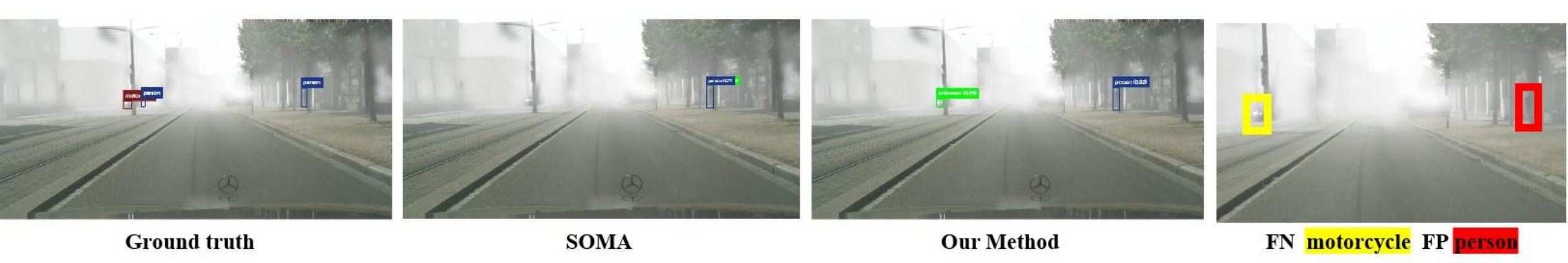}\vspace{-0.5em}
	{\footnotesize  The motorcycle in the fog has not been detected, while the person has been mistakenly labeled as novel category.}\vspace{0.5em}
	\includegraphics[width=\textwidth]{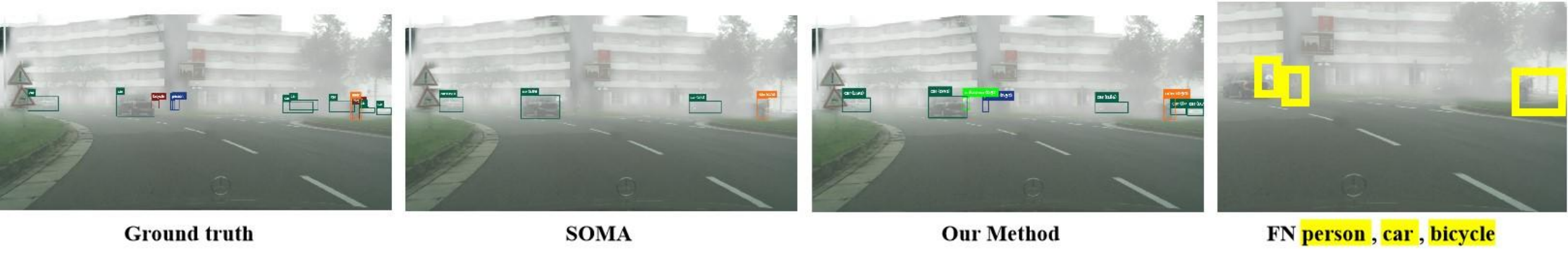}\vspace{-0.5em}
	{\footnotesize The person, car, and bicycle in the fog have all gone undetected.}\vspace{0.5em}
	\includegraphics[width=\textwidth]{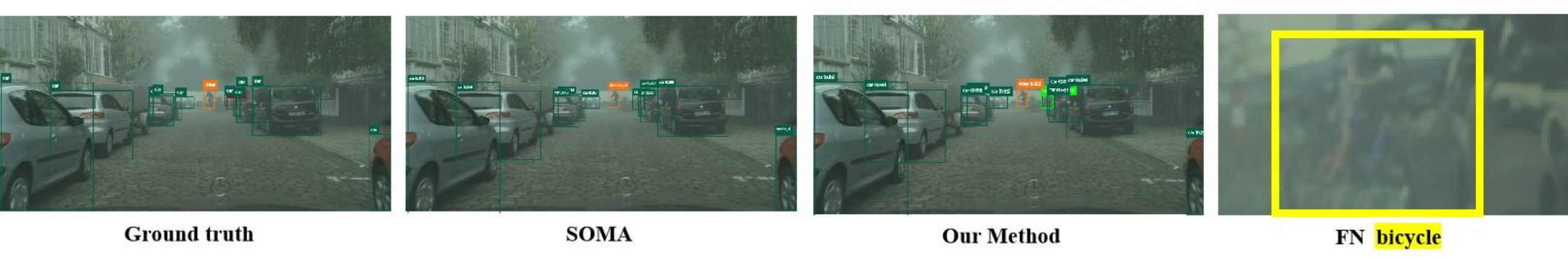}\vspace{-0.5em}
	{\footnotesize The bicycle concealed within the cars has not been detected.}\vspace{0.5em}
	\includegraphics[width=\textwidth]{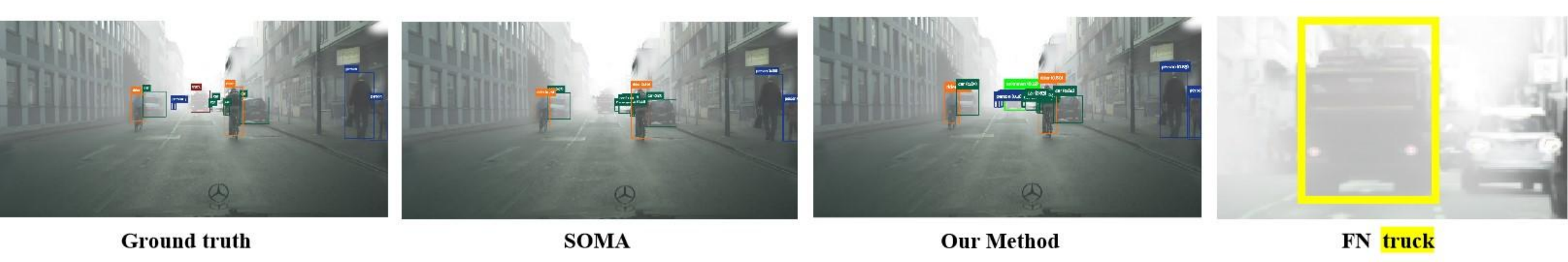}\vspace{-0.5em}
	{\footnotesize The truck at the end of the road cannot be detected.}\vspace{0.5em}
	\caption{Visualization of detection results in Cityscapes\cite{cordts2016cityscapes} $\rightarrow$ Foggy Cityscapes \cite{sakaridis2018semantic} dataset. The first column shows the groundtruth; the second and third columns visualize the detection results of SOMA \cite{li2023novel} and CCKM, respectively. In the last column, the closeups of false positive (FP) and false negative (FN) errors are zoomed in \textcolor{red}{red} and \textcolor{yellow}{yellow} boxes. The detection errors of SOMA are described in the captions below each row.}
	\label{fig: det city}
	\vspace{-0.5em}
\end{figure*}

\subsubsection{Connection Matrix Analysis}
This ablation study investigates the impact of two types of connection matrices: connection matrix of source domain (SCM, $U_s$) and connection matrix of target domain (TCM, $U_t$). The experiments are conducted without incorporating additional components (CMB or AFA) under the het-sem setting with 5 novel classes on the Cityscapes $\rightarrow$ Foggy Cityscapes task in Table \ref{tab:ablation2} . SCM only ($U_s$) leads to a higher mAP (46.65), suggesting improved localization for base classes due to better source feature correlation. Consistent with Table \ref{tab:protoball} , using ProtoBall distance to construct SCM reduces overlap between base and novel feature distributions, thereby alleviating base–novel confusion. TCM only ($U_t$) achieves the best AR (3.31), emphasizing its strength in retrieving novel-class objects by leveraging target-domain feature topology. It also lowers AOSE to 824, outperforming SCM alone in open-set filtering. Combining both connection matrices yields the best overall results.

\subsubsection{Parameter Analysis}
We conduct sensitivity studies for $\gamma$, $\beta$ and $K$ under the het-sem setting on the Cityscapes $\rightarrow$ Foggy Cityscapes benchmark with 5 novel classes. As shown in Fig.~\ref{fig:parameter analysis}, a moderate $\gamma$ effectively enlarges the inter-class margin, helping distinguish novel categories from the background while reducing feature overlap with base categories. However, an excessively large $\gamma$ may misclassify unmatched object query features belonging to novel categories as background, leading to the observed drop in AR. Regarding the momentum parameter $\beta$, Table~\ref{tab:beta_sensitivity} shows that the performance varies smoothly within a reasonable range, and the best overall balance is achieved at $\beta = 1\mathrm{e}{-2}$. As shown in Table~\ref{tab:k_sensitivity}, $K = 3$ slightly improves mAP but increases WI and AOSE due to a compact yet incomplete novel-class prototype region that weakens open-set discrimination. When $K = 7$, less representative candidates are introduced, degrading prototype purity and increasing WI and AOSE. Overall, $K = 5$ yields the best trade-off.

\subsection{Qualitative Analysis} \label{sec:visual}

\noindent\textbf{t-SNE visualization of distance metrics.}
We presents a t-SNE visualization of object query features on Cityscapes $\rightarrow$ Foggy Cityscapes under the freq-dec setting with three novel classes. As shown in Fig. \textcolor{red}{\ref{fig:tsne}}. (a), when using cosine distance as the metric, object query features for novel categories can be partially separated from the background. However, they still exhibit noticeable overlap with object query features for base categories, indicating a bias toward specific base class in the feature space. In contrast, Fig. \textcolor{red}{\ref{fig:tsne}}. (b) illustrates the results obtained with the proposed ProtoBall distance. Although object query features for novel categories occupy a relatively larger region due to the presence of multiple novel classes, their overlap with base categories is substantially reduced. This observation suggests that ProtoBall distance effectively mitigates the attraction of novel features toward individual base class prototypes, while preserving sufficient separability from the background.

\noindent\textbf{Visualization of detection results.}
Samples from  Cityscapes $\rightarrow$ Foggy Cityscapes are selected for comparison with SOMA \cite{li2022sigma}. The detection results are presented in Fig. \textcolor{red}{\ref{fig: det city}}. Under foggy conditions, SOMA fails to detect key objects such as the motorcycle, car, person, and bicycle. These objects are partially occluded or appear with reduced contrast, indicating that SOMA struggles with degraded visual inputs and context understanding. As for false novel predictions, SOMA incorrectly labels a person as a novel category, highlighting limitations in semantic discrimination. This misclassification suggests that SOMA's feature representation may lack robustness when encountering domain-shifted or visually ambiguous instances. As for object occlusion handling, the bicycle obscured by surrounding cars is not detected by SOMA, implying inadequate performance under partial occlusion. Similarly, the truck at the end of the road, which appears distant and partially covered by fog, is completely missed.

\section{Discussion and Conclusion} \label{sec:discuss}
This paper presents a new adaptive open-set object detection (AOOD) framework grounded in category-level knowledge mining. Specifically, clustering-based memory bank is first constructed to store both ategory-level knowledge across domains. The memory bank is iteratively updated through unsupervised clustering, which facilitates the mining of discriminative category-level features. To effectively handle novel categories, a base-to-novel selection metric is introduced to identify high-quality feature representations of novel classes in the source domain. The selection process is guided by the category-level knowledge of base categories in the memory bank. These selected features are subsequently used to refine and enhance the memory bank. Furthermore, an adaptive feature assignment strategy is proposed to assign category labels to features based on the memory bank. All features assigned with category labels are incorporated to further reinforce the category-level knowledge stored in the memory bank.  


Future work will focus on extending this framework by exploring how to effectively distill category-level knowledge, aiming to bridge the semantic gap between coarse-grained category representations and fine-grained individual features.

\bibliographystyle{IEEEtran}
\bibliography{refs.bib}

\begin{IEEEbiography}[{\includegraphics[width=1in,height=1.25in,clip,keepaspectratio]{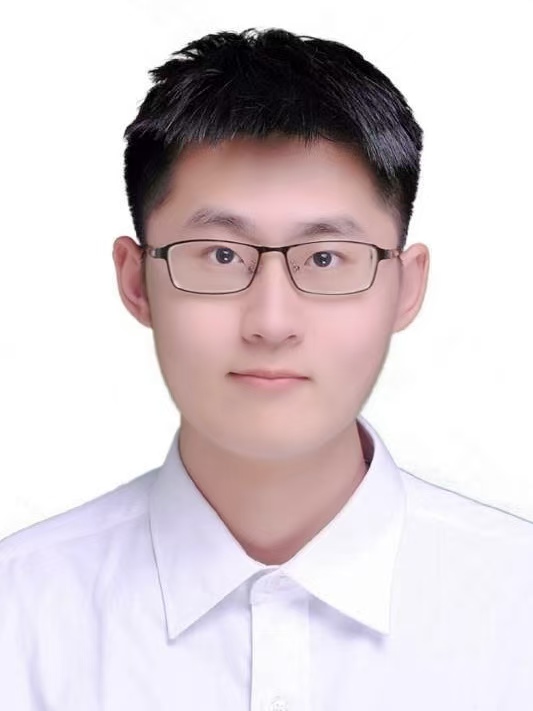}}]{Yuqi Ji} received the B.Sc. degree in Detection, Guidance and Control Technology in 2022 from Xidian University, Xi'an, China, where he is currently working toward the Ph.D. degree. His research interests include object detection and computer vision.
\end{IEEEbiography}

\begin{IEEEbiography}[{\includegraphics[width=1in,height=1.25in,clip,keepaspectratio]{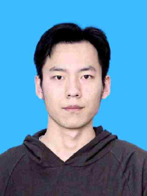}}]{Junjie Ke} is currently a postdoctoral researcher at the School of Software, Tsinghua University. He received his Ph.D. degree in Circuits and Systems from Xidian University, Xi'an, China, in 2025. He also obtained his B.Sc. degree in Electronic and Information Engineering from Xidian University in 2019. His research interests focus on object detection and computer vision.
\end{IEEEbiography}

\begin{IEEEbiography}[{\includegraphics[width=1in,height=1.25in,clip,keepaspectratio]{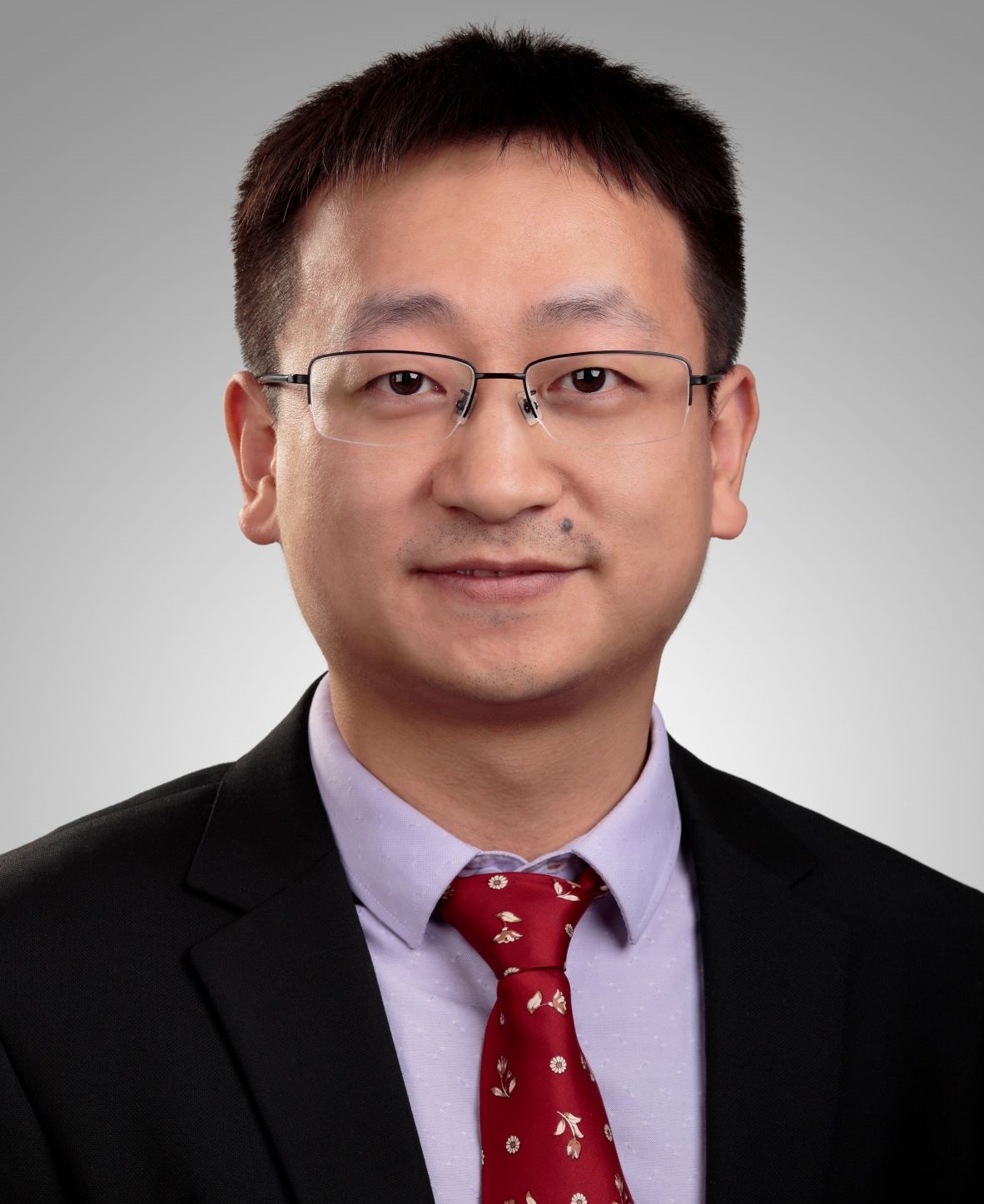}}]{Lihuo He} (Member, IEEE) received the B.Sc. degree in electronic and information engineering and the Ph.D. degree in pattern recognition and intelligent systems from Xidian University, China, in 2008 and 2013, respectively. He is currently a Professor in the School of Electronic Engineering at Xidian University. His research interests include image/video quality assessment, cognitive computing, and computational vision. In these areas, he has published several scientific articles in refereed journals including the IEEE TPAMI, TIP, TMM, TCYB and TCSVT, and conferences including the CVPR, IJCAI and AAAI.
\end{IEEEbiography}

\begin{IEEEbiography}[{\includegraphics[width=1in,height=1.25in,clip,keepaspectratio]{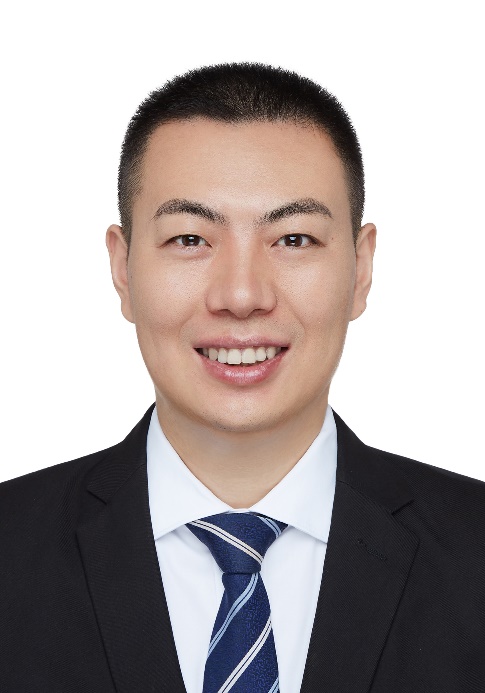}}]{Lizhi Wang} (Member, IEEE) received the BS and PhD degrees from Xidian University, Xi’an, China, in 2011 and 2016, respectively. He is currently a professor with the School of Artificial Intelligence, Beijing Normal University. His research interests include computational photography and image processing. He is serving as an associate editor of IEEE Transactions on Image Processing. He received the Best Paper Runner-up Award of ACM MM 2022 and Best Paper Award of IEEE VCIP 2016.
\end{IEEEbiography}

\begin{IEEEbiography}[{\includegraphics[width=1in,height=1.25in,clip,keepaspectratio]{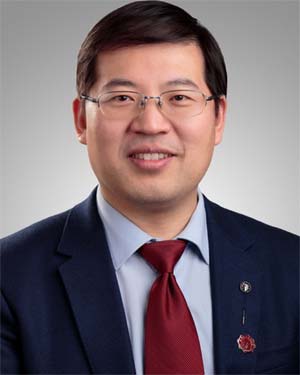}}]{Xinbo Gao} (Fellow, IEEE) received the B.Eng., M.Sc., and Ph.D. degrees in electronic engineering, signal and information processing from Xidian University, Xi’an, China, in 1994, 1997, and 1999, respectively. From 1997 to 1998, he was a Research Fellow with the Department of Computer Science, Shizuoka University, Shizuoka, Japan. From 2000 to 2001, he was a Postdoctoral Research Fellow with the Department of Information Engineering, The Chinese University of Hong Kong, Hong Kong. Since 2001, he has been with the School of Electronic Engineering, Xidian University. He is also a Cheung Kong Professor with the Ministry of Education of China, Professor of pattern recognition and intelligent system with Xidian University, and Professor of computer science and technology with the Chongqing University of Posts and Telecommunications, Chongqing, China. He has authored or coauthored seven books and around 300 technical articles in refereed journals and proceedings. His current research interests include image processing, computer vision, multimedia analysis, machine learning, and pattern recognition. He was the General Chair or Co-Chair, Program Committee Chair or Co-Chair or PC Member for around 30 major international conferences. He is on the Editorial Boards of several journals, including Signal Processing (Elsevier) and Neurocomputing (Elsevier). He is a fellow of IET, AAIA, CIE, CCF, and CAAI.
\end{IEEEbiography}
\end{document}